\documentclass{article} 
\usepackage{iclr2027_conference,times}
\usepackage{mathptmx}
\DeclareSymbolFont{cmsy}{OMS}{cmsy}{m}{n}
\DeclareSymbolFontAlphabet{\mathcal}{cmsy}

\usepackage{amsmath,amsfonts,bm}

\def\eqref#1{equation~\ref{#1}}

\def\1{\bm{1}}

\DeclareMathAlphabet{\mathsfit}{\encodingdefault}{\sfdefault}{m}{sl}
\SetMathAlphabet{\mathsfit}{bold}{\encodingdefault}{\sfdefault}{bx}{n}

\usepackage{hyperref}
\usepackage{url}
\hypersetup{
  hidelinks,
  pdftitle={Vision-Language-Action or World-Action? Just Action Models},
  pdfauthor={Yonghao Yu, Lang Huang, Runyi Li, Toshihiko Yamasaki}
}
\makeatletter
\def\NAT@hyper@#1{\mbox{\hyper@natlinkstart{\@citeb\@extra@b@citeb}#1\hyper@natlinkend}}
\makeatother
\usepackage{booktabs}
\usepackage{array}
\usepackage{graphicx}
\usepackage{amsmath}
\usepackage{amssymb}
\usepackage{multirow}
\usepackage{xcolor}
\usepackage{tikz}
\usepackage{placeins}
\usepackage{pifont}
\usetikzlibrary{arrows.meta,fit,backgrounds,positioning,calc,shapes.geometric,patterns}

\newcommand{\modelname}{{JAM}}
\newcommand{\modelfullname}{Just Action Model}

\title{
Vision-Language-Action or World-Action?\\
Just Action Models
}

\newcommand{\jamauthor}[3]{%
  \parbox[t]{\dimexpr\textwidth/2-2\tabcolsep\relax}{%
    \normalfont\raggedright
    \textbf{#1}\\
    #2\\
    \texttt{#3}%
  }%
}
\author{%
\jamauthor{Yonghao Yu}{The University of Tokyo}{\href{mailto:y_yu@cvm.t.u-tokyo.ac.jp}{y\_yu@cvm.t.u-tokyo.ac.jp}}%
\And
\jamauthor{Lang Huang\thanks{Corresponding author.}}{National Institute of Informatics}{\href{mailto:lang@nii.ac.jp}{lang@nii.ac.jp}}%
\AND
\jamauthor{Runyi Li}{The University of Tokyo}{\href{mailto:r_li@cvm.t.u-tokyo.ac.jp}{r\_li@cvm.t.u-tokyo.ac.jp}}%
\And
\jamauthor{Toshihiko Yamasaki}{The University of Tokyo}{\href{mailto:yamasaki@cvm.t.u-tokyo.ac.jp}{yamasaki@cvm.t.u-tokyo.ac.jp}}%
}

\iclrfinalcopy

\begin{document}

\maketitle
\fancyhead{}
\renewcommand{\headrulewidth}{0pt}

\begin{abstract}
The vision-language-action (VLA) and world-action model (WAM) communities debate whether control policies should inherit pretrained knowledge from vision-language models or from visual generators. The debate, however, bundles three separable choices: backbone initialization, the objective applied to future observations, and whether the machinery for that future loss is kept at inference. A gap between a VLA and a WAM therefore cannot simply be attributed to any one of them. We show that decoupling these choices yields a common framework for both families, which we term \emph{\modelfullname{}} (\modelname{}): a pretrained visual backbone with an action expert, trained with an action loss and a foresight loss. The foresight loss trains current-observation features to predict frozen encodings of the recorded future through a prediction head discarded after training, so no visual generation is involved in either training or inference, and the backbone can be chosen freely.
We first compare objectives within fixed policy architectures, then evaluate the foresight recipe across backbone families.
The resulting \modelname{} policies outperform their matched FastWAM and ImageWAM counterparts on LIBERO and LIBERO-Plus, and achieve similar or slightly higher success rates on RoboTwin. Compared with FastWAM, \modelname{} improves the success rates from 97.0\%/55.3\%/91.2\% to 98.2\%/64.5\%/91.4\%, respectively. With cached foresight targets, JAM reduces estimated online training FLOPs per sample per update by 73\% relative to FastWAM.
Our results show that future supervision improves control without visual generation or additional inference machinery, and strong foresight performance is attainable with either generator-initialized or vision-language backbones.
\textbf{Project Page:} \url{https://y0uroy.github.io/JAM}
\end{abstract}

\section{Introduction}

\begin{figure}[t]
\centering
\includegraphics[width=\textwidth]{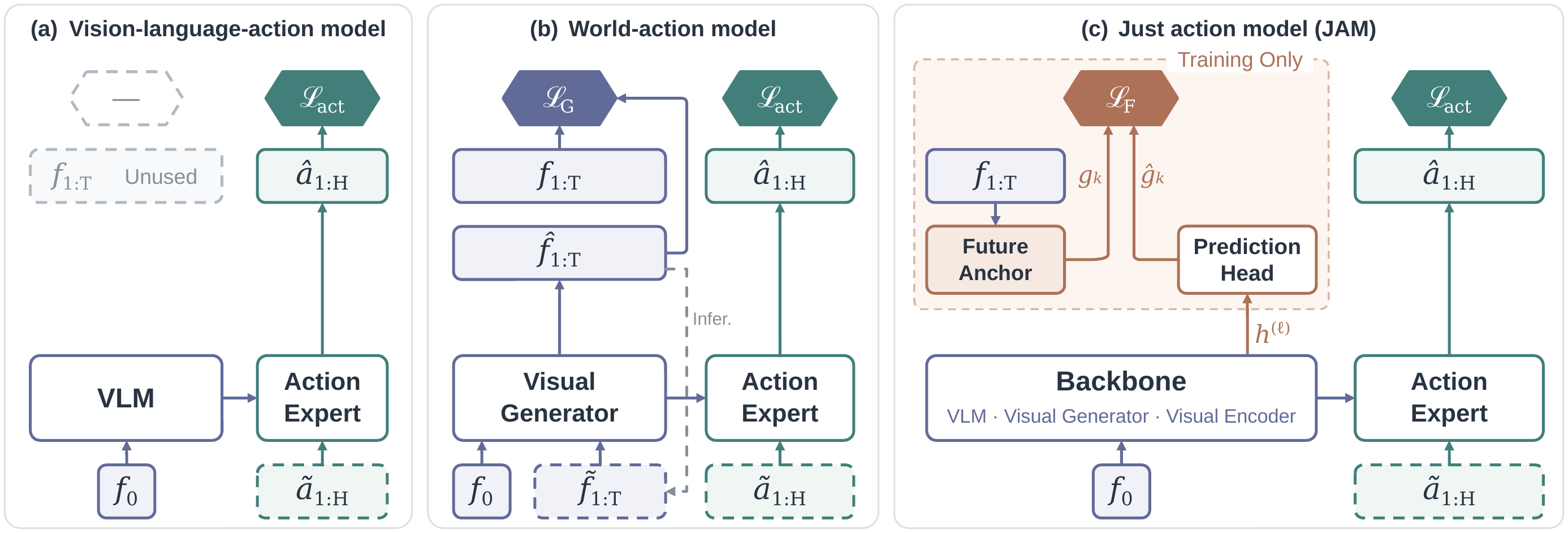}
\vspace{-3mm}
\caption{The three choices that separate VLAs, WAMs, and \modelname{}: backbone initialization, the objective on recorded future observations, and the modules kept at inference. (a) The VLA reference uses no future objective, only the action loss $\mathcal{L}_{\mathrm{act}}$. (b) WAMs add visual generation loss $\mathcal{L}_{\mathrm G}$. The dashed loop shows generated future observations entering the action expert in some WAMs. (c) \modelname{} uses foresight loss $\mathcal{L}_{\mathrm F}$ across pretraining families, discards its supervision branch, and retains the corresponding action path.}
\label{fig:designspace}
\vspace{-5mm}
\end{figure}

Pretrained visual models provide a starting point for robot control. Standard vision-language-action models (VLAs) adapt vision-language representations to predict actions \citep{openvla2024,pi05_2025,pi0_2024}. World-action models (WAMs) start from image or video generators and learn actions together with future visual prediction \citep{dreamzero2026,fastwam2026,lingbotva2026,motus2026,imagewam2026}. These approaches differ, as depicted in Figure~\ref{fig:designspace}, in both their pretrained knowledge and how they use the robot trajectories: most VLAs learn only the actions in a trajectory, whereas a WAM also learns to generate the observations that follow. Comparing their success rates alone does not identify which difference matters.

We separate three choices: \emph{backbone initialization}, \emph{the objective on recorded future observations}, and \emph{the modules used at inference}. Recent attempts such as FastWAM~\citep{fastwam2026} and ImageWAM~\citep{imagewam2026} remove test-time video or image generation and suggest that visual generative objectives at training time are the key to WAMs. We ask a more basic question: \emph{does learning from recorded future observations require a visual generative objective at all? If not, can backbones without visual generative pretraining learn from future observations in the same way?}


We show that predicting future representations can replace visual generative supervision in policy training.
We introduce \emph{\modelfullname{}} (\modelname{}), a common recipe for a pretrained visual backbone and an action expert. The policy is trained with an action loss and a foresight loss: a prediction head maps current-observation features to targets supplied by a frozen future anchor. Future observations provide targets for the foresight loss, never inputs to the action path. The anchor branch and prediction head are used only during training, and fixed targets can be cached. At deployment, the backbone encodes the current context and the action expert predicts the action chunk. \modelname{} requires no image or video generation during policy fine-tuning or deployment. The backbone may still come from generative pretraining, and the action expert still uses flow matching.

We evaluate the recipe in two stages. First, we replace the generative objectives of FastWAM and ImageWAM while keeping each policy's architecture, data, optimizer, training budget, and deployment path fixed. Predicting future representations preserves or improves success, and targets at future positions are more effective than those at the current position. Backbone and future anchor pretraining play different roles: a pretrained FastWAM backbone remains effective with a randomly initialized, frozen future anchor, and direct VAE features give its strongest result. \modelname{} (Ovis-U1)  performs best with targets from the pretrained Yak backbone, so the preferred target is architecture-dependent. With the best target for each architecture, LIBERO-Plus success rises from $55.31\%$ to $64.54\%$ for FastWAM and from $65.27\%$ to $79.95\%$ for ImageWAM. With cached targets, this recipe reduces estimated online FLOPs per sample per update by $73.0\%$ for FastWAM and $36.8\%$ for ImageWAM.

Second, we retain the common foresight objective and instantiate it with generative, self-supervised, and vision-language backbones. On LIBERO-Plus, \modelname{} (Ovis-U1), \modelname{} (DINOv2-L), and \modelname{} (Qwen3.5-2B) reach $79.95\%$, $70.39\%$, and $81.14\%$, against $65.27\%$ for ImageWAM (Ovis-U1), which is trained with the generation loss $\mathcal{L}_{\mathrm G}$. These results establish the recipe's applicability across all three families: generative backbones can learn from future observations without a visual generative loss, and the same supervision applies to non-generative backbones.

Our contributions are:

\begin{itemize}
\item We propose \modelname{}, a single recipe for a pretrained visual backbone and an action expert in which recorded future observations enter only as frozen prediction targets, and the foresight branch is discarded after training.
\item We show through fixed-architecture comparisons that foresight can replace visual generation while preserving or improving success. Future-offset targets are more effective than current-offset targets, and the best target depends on the backbone: a plain VAE suffices for the Wan video backbone. With targets cached, JAM reduces the estimated online training FLOPs per sample per update by $73.0\%$ relative to FastWAM.
\item We apply the same objective to generative, self-supervised, and vision-language backbones, with the strongest generative and vision-language instantiations achieving comparable control performance, showing that strong performance with foresight is attainable using either a generator-initialized or a vision-language backbone.
\end{itemize}

\section{Related Work}

\paragraph{Pretrained models for robot control.}
OpenVLA adapts a vision-language model to autoregressive action prediction; $\pi_0$ and $\pi_{0.5}$ pair vision-language backbones with flow-matching action experts \citep{openvla2024,pi05_2025,pi0_2024}. WAMs combine generator initialization with future visual prediction \citep{gigaworld2026,dreamzero2026,lingbotva2026,motus2026}. FastWAM retains video co-training without test-time video generation~\citep{fastwam2026}; ImageWAM uses an image-editing backbone and a future-image target~\citep{imagewam2026}. \modelname{} replaces their visual generation objectives while preserving each policy's architecture and deployment path, separating the use of pretrained generators from the need to generate during policy training.

\paragraph{Future-representation supervision.}
Future observations support planning and predictive supervision \citep{gr1_2024,gr2_2024,pad2024,uva2025}. Joint-embedding methods avoid pixel reconstruction \citep{vjepa2_2025,tdv2026}; Var-JEPA studies a variational formulation on tabular data~\citep{varjepa2026}. Mirai and Video-Mirai apply foresight to image and video generation \citep{mirai2026,videomirai2026}. For control, FRAPPE aligns policy features with multiple future representations, VLAFlow pairs latent alignment with language-supervised co-training, and Being-H0.7 aligns a deployable prior with a future-informed posterior \citep{frappe2026,vlaflow2026,beingh07_2026}. VLA-JEPA uses frozen video targets for human-video pretraining and joint representation and action learning~\citep{vlajepa2026}; PFD distills future-conditioned action residuals into a current-only FastWAM policy while retaining video co-training and a deployed residual adapter~\citep{pfd2026}.

\paragraph{Predictive supervision for action policies.}
FLARE mixes future, state, and action tokens through self-attention in the action DiT~\citep{flare2025}. JEPA-WAM supervises a shared action-generating predictor using dense joint current--future targets in a frozen V-JEPA space. Its pretrained-VLA instantiation preserves the original pathways and removes the prediction branch at deployment~\citep{jepawam2026}.
\modelname{} differs from FLARE in where prediction is applied: a separate training-only head supervises existing visual-backbone tokens, preserving the matched policy's action computation. In contrast to JEPA-WAM's joint current--future V-JEPA target, it supervises future-position features in a target space chosen for each backbone. This common recipe replaces visual generation in existing WAMs and extends across generative, self-supervised, and vision-language backbones. Appendix~\ref{app:extendedrw} gives further comparisons and more related work.

\section{JAM: Just Action Models}
\label{sec:method}

\subsection{Problem formulation}
\label{sec:formulation}

All three recipes in Figure~\ref{fig:designspace} share the same data and the same deployed interface. A training example contains the current observation $f_0$, the robot state $s_0$, an instruction $c$, an action chunk $a_{1:H}$, and recorded future observations $f_{1:T}$. The deployed policy predicts $a_{1:H}$ from $f_0$, $s_0$, and $c$. The recipes differ in which pretrained model initializes the backbone, and what loss is placed on the recorded futures $f_{1:T}$.

\begin{figure}[t]
\centering
\includegraphics[width=\textwidth]{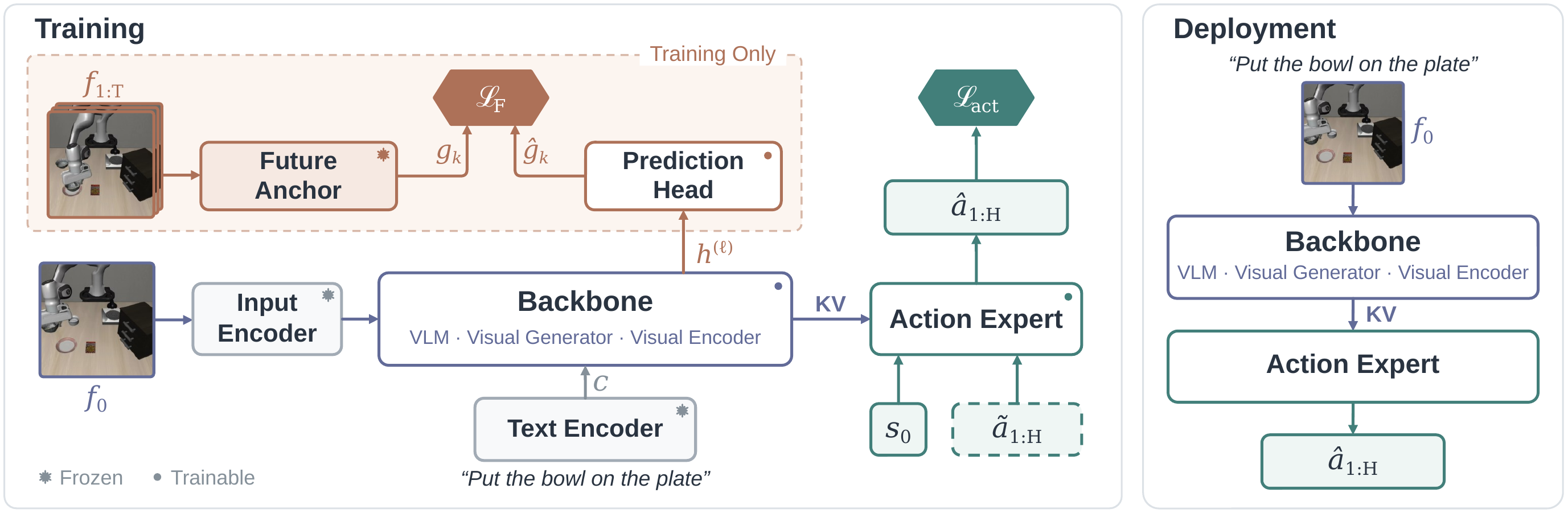}
\vspace{-7mm}
\caption{\modelname{} training (left) and deployment (right), with the notation of Section~\ref{sec:alignment}. The input and text encoders stand for architecture-specific interfaces. Recorded futures $f_{1:T}$ enter only the future anchor branch inside the dashed box; deployment discards this branch and the prediction head, leaving the backbone and action expert.}
\label{fig:method}
\vspace{-2mm}
\end{figure}

The action-only VLA reference adapts a pretrained vision-language backbone and an action expert with the action flow-matching loss alone~\citep{flowmatching2022},
\vspace{-2mm}
\begin{equation}
\mathcal{L}_{\mathrm{VLA}}=\mathcal{L}_{\mathrm{act}},
\label{eq:vla}
\vspace{-2mm}
\end{equation}
where $\mathcal{L}_{\mathrm{act}}$ supervises the action chunk. The recorded futures are unused, and pretrained weights enter through policy initialization. The WAM reference starts from a pretrained generator and adds its generative loss $\mathcal{L}_{\mathrm G}$ on the recorded futures,
\vspace{-2mm}
\begin{equation}
\mathcal{L}_{\mathrm{WAM}}=\mathcal{L}_{\mathrm{act}}+\lambda_{\mathrm G}\,\mathcal{L}_{\mathrm G}.
\label{eq:wam}
\vspace{-2mm}
\end{equation}
This term ties the recipe to a generator: it needs a decoder or denoising path during training, and in some WAMs also at inference.
\modelname{} pairs a backbone with an action expert and optimizes
\vspace{-2mm}
\begin{equation}
\mathcal{L}_{\mathrm{JAM}}=\mathcal{L}_{\mathrm{act}}+\lambda_{\mathrm F}\,\mathcal{L}_{\mathrm F},
\label{eq:total}
\vspace{-2mm}
\end{equation}
as shown in Figure~\ref{fig:method}. The foresight loss $\mathcal{L}_{\mathrm F}$ trains current-observation features to predict the future anchor's representation of the recorded futures, without a visual generative loss. The anchor branch and prediction head are used only during training; the deployed architecture is unchanged, and no future frame is ever generated, either in training or at inference.

Equation~\ref{eq:total} bridges the two families: the futures supervise through the frozen future anchor rather than through visual generation. The objective does not require an image or video generator, so the backbone can be a generative DiT, a self-supervised encoder, or a vision-language model. The future anchor can use a frozen backbone or reuse the policy's VAE tokenizer. The deployed policy consists of the backbone and the action expert.

\subsection{Foresight supervision}
\label{sec:alignment}

\paragraph{Policy and target features.}
The visual backbone encodes $f_0$, conditioned on $c$ when the backbone supports text input. The action expert receives the resulting visual context, $c$, the robot state $s_0$, and the noised action chunk $\widetilde a_{1:H}$ at a sampled flow time. Let $\mathbf h^{(\ell)}\in\mathbb{R}^{N\times D}$ denote the current-observation tokens at backbone block $\ell$. The frozen future anchor $E_\varphi$ processes the recorded futures $f_{1:T}$ through its own conditioning interface. For Yak, FLUX.2, and Qwen, this interface also reads $f_0$ and $c$. The target $\mathbf g_k\in\mathbb{R}^{N\times D_g}$ contains the anchor's features at readout offset $k$, its $k$-th temporal position, where $k=0$ is the current observation and $k\geq1$ are recorded futures. Backbone features or VAE latents are matched to the spatial-token layout of $\mathbf h^{(\ell)}$, and $D_g$ is the target feature dimension. With fixed target preprocessing and conditioning, these targets can be computed once and cached.

\paragraph{Foresight loss.}
The trainable prediction head $p$ maps the policy tokens $\mathbf h^{(\ell)}$ into the target space, giving one prediction $\hat{\mathbf g}_k\in\mathbb{R}^{N\times D_g}$ per offset $k$. The foresight loss is the cosine distance between prediction and target, averaged over tokens and offsets:
\vspace{-1mm}
\begin{equation}
\mathcal{L}_{\mathrm F}=\frac{1}{|\mathcal K|}\sum_{k\in\mathcal K}\mathbb{E}_{n}\Big[1-\cos\big(\hat{\mathbf g}_{k,n},\ \mathrm{sg}[\mathbf g_{k,n}]\big)\Big],
\label{eq:fra}
\vspace{-1mm}
\end{equation}
where $\mathcal K$ is the set of supervised readout offsets and $n$ indexes spatial tokens. The operator $\mathrm{sg}$ stops gradients through the target. Gradients reach the head and the policy backbone, so recorded futures supervise the features read by the action expert without entering the action path. Supervision is applied at the selected current or future target positions. The head is a small token-mixing transformer with one linear readout per offset.

\paragraph{Action inference.}
The foresight branch exists only in training. After training, we discard the anchor branch and prediction head, and neither target extraction nor future prediction runs at inference. The policy encodes $f_0$ and $c$ once, and the action expert reuses this context across its flow-matching sampling steps. This is the inference procedure of the matched FastWAM and ImageWAM baselines, which already avoid test-time future generation. The difference is that \modelname{} avoids visual generation during training as well.

\subsection{Instantiations across pretraining families}
\label{sec:instantiations}

We instantiate \modelname{} with backbones from three pretraining families. In the cross-backbone comparison in Table~\ref{tab:imagewamarchmatched}, each future anchor is a frozen copy of the corresponding backbone. Our matched FastWAM and ImageWAM experiments additionally evaluate VAE-direct future anchors.

\paragraph{Generative backbones.}
FastWAM pairs a Wan2.2-5B video backbone with an action expert of the same depth, and ImageWAM (Ovis-U1) pairs the pretrained Yak18 image-editing backbone with an 18-block action expert of matching width \citep{wan2025,fastwam2026,imagewam2026}. Their generative objectives denoise future video latents and generate a future image. Under \modelname{}, the backbone sees only the clean current frame, and the policy architecture and action-expert initialization are unchanged. Two further targets isolate what the frozen generator contributes: the same backbone with random initialization, which keeps the architecture but removes the pretrained features, and the frozen VAE tokenizer alone, which removes the target transformer and predicts lifted latent-patch features.

\paragraph{Self-supervised backbones.}
DINOv2 is an image encoder with no language input and no generative head \citep{dinov2_2024}. The action expert mirrors the backbone's transformer blocks, and a separate text encoder supplies instruction features to the action expert, but not to the visual backbone.

\paragraph{Vision-language backbone.}
Qwen3.5-2B is a vision-language model that already reads images and text together \citep{qwen35_2026}. The action expert mirrors the backbone's blocks and reads the backbone's tokens through causal token mixing over image, text, and action tokens, so it is a VLA trained with an additional foresight loss.

\section{Experiments}

Our study has two stages. First, we replace visual generation with foresight within fixed policy architectures and deployment paths, examining target offsets and future anchors. Second, we apply the common objective to backbones from different pretraining families. We also compare the resulting policies with published systems and evaluate training cost.
\subsection{Setup}
\label{sec:setup}

\paragraph{Benchmarks.}
LIBERO contains the Spatial, Object, Goal, and Long suites \citep{libero2023}. We evaluate 40 tasks with 50 rollouts each and report success per suite and averaged across suites. For LIBERO-Plus, we follow ImageWAM's $15\%$ subset protocol \citep{liberoplus2025,imagewam2026}: a seeded draw per suite produces a fixed manifest of $1{,}506$ perturbed tasks, evaluated once each and shared by all local policies. RoboTwin 2.0 evaluates 50 bimanual tasks under Clean and Randomized settings \citep{robotwin2_2025}.

\paragraph{Training and comparisons.}
Each generative policy and its foresight counterpart share data, policy architecture, optimizer, training budget, and evaluator. Training budgets follow FastWAM and ImageWAM \citep{fastwam2026,imagewam2026}. On LIBERO, the FastWAM recipe runs $20$k updates and the ImageWAM recipe, used by the other backbones, $21.7$k, both at global batch 128. On RoboTwin 2.0, both recipes run five epochs, $29{,}355$ updates at global batch $1{,}024$ for FastWAM and $69{,}900$ at global batch 384 for ImageWAM.

\paragraph{Foresight configurations.}
For anchors that use a backbone, the policy and anchor readouts are at about $60\%$ of their depth: block 18 of 30 for FastWAM, Yak layer 11 of 18 for ImageWAM, and layer 13 of 24 for Qwen3.5. DINOv2 is the exception and is supervised at its final block. Foresight losses are averaged over readout offsets, with total weight $\lambda_{\mathrm F}=0.6$ for FastWAM and $0.3$ for the other policies. Readout indices count positions in the anchor's temporal sequence. FastWAM samples frames every four environment steps, and its causal VAE encodes the current frame alone and merges each following group of four frames into one latent position. Offset $[0]$ is therefore the current frame, and offsets $[1]$ and $[2]$ cover steps 4 to 16 and 20 to 32, respectively. For the other backbones, the recorded future is a single frame $16$ steps ahead, so readout offset $[1]$ denotes that frame and $[0]$ the current one. More training details are given in Appendix~\ref{app:impl}.

\subsection{Replacing generation with foresight}
\label{sec:objective}

\begin{figure}[t]
\centering
\definecolor{wgreen}{HTML}{427F7A}\definecolor{wink}{HTML}{293340}\definecolor{wmuted}{HTML}{7A8590}
\definecolor{wpanel}{HTML}{C8D4DE}
\resizebox{\textwidth}{!}{%
\begin{tikzpicture}[x=1cm,y=1cm]
  \tikzset{
    tasktitle/.style={font=\scriptsize\bfseries,text=wink,anchor=south,inner sep=0pt},
    stage/.style={font=\scriptsize,text=wmuted,anchor=base,inner sep=0pt},
    method/.style={font=\scriptsize\bfseries,text=wink,rotate=90,anchor=center,inner sep=0pt},
  }
  \newcommand{\rollframe}[3]{\node[anchor=south west,inner sep=0pt] at (#1,#2) {\includegraphics[width=1.6cm]{#3}};}
  \node[method] at (0.20,2.365) {ImageWAM};
  \node[method] at (0.20,0.825) {\modelname{}};

  \node[tasktitle] at (3.755,3.46) {(a) Put the black bowl in the bottom drawer and close it};
  \foreach \k/\name in {0/Initial,1/Grasp,2/Place,3/Terminal}{
    \node[stage] at ({0.42+\k*1.69+0.8},3.20) {\name};
    \rollframe{0.42+\k*1.69}{1.64}{Figures/qualitative/drawer_native_\k.png}
    \rollframe{0.42+\k*1.69}{0.10}{Figures/qualitative/drawer_mirai_\k.png}
  }

  \draw[wpanel,line width=0.6pt] (7.27,0.10)--(7.27,3.82);

  \node[tasktitle] at (10.785,3.46) {(b) Put both moka pots on the stove};
  \foreach \k/\name in {0/Initial,1/First pot,2/Second pot,3/Terminal}{
    \node[stage] at ({7.45+\k*1.69+0.8},3.20) {\name};
    \rollframe{7.45+\k*1.69}{1.64}{Figures/qualitative/moka_native_\k.png}
    \rollframe{7.45+\k*1.69}{0.10}{Figures/qualitative/moka_mirai_\k.png}
  }
\end{tikzpicture}%
}
\vspace{-7mm}
\caption{Matched LIBERO-Long rollouts for ImageWAM and \modelname{}, both on Ovis-U1.}
\label{fig:pairedrollouts}
\vspace{-3mm}
\end{figure}
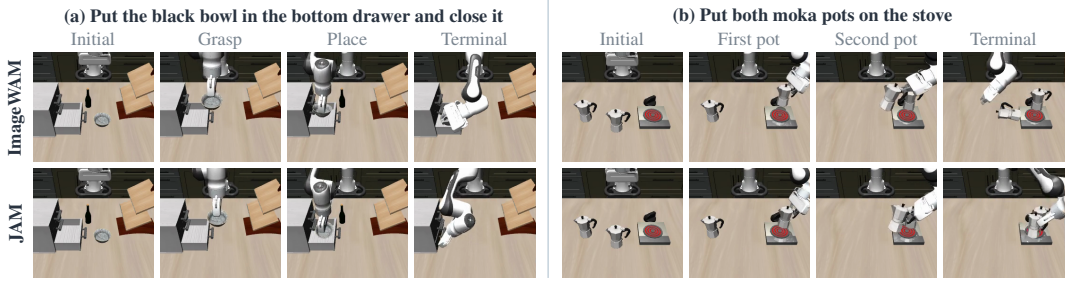

\begin{table}[t]
\vspace{-2mm}
\caption{Generation versus foresight supervision with the policy architecture fixed. $\mathcal{L}_{\mathrm G}$ is the generative loss and $\mathcal{L}_{\mathrm F}$ the foresight loss. The target rows also vary the future anchor. Readout offsets are the set $\mathcal{K}$ of target positions, with $0$ the current frame. The total foresight weight $\lambda_{\mathrm F}$ is fixed at $0.6$ for Wan and $0.3$ for Ovis-U1 and is divided equally among the readout offsets.}
\vspace{1.5mm}
\label{tab:objectives}
\centering
\small\setlength{\tabcolsep}{5pt}
\renewcommand{\arraystretch}{1.25}\setlength{\aboverulesep}{0pt}\setlength{\belowrulesep}{0pt}
\begin{tabular}{lcccccc}
\toprule
Policy & $\mathcal{L}_{\mathrm G}$ & $\mathcal{L}_{\mathrm F}$ & Future anchor & Readout offsets & LIBERO & LIBERO-Plus \\
\midrule
\multirow{2}{*}{FastWAM} & $\times$ & $\times$ & -- & -- & 95.25 & 45.55 \\
 & \checkmark & $\times$ & -- & -- & 97.00 & 55.31 \\
\cmidrule{1-7}
\multirow{5}{*}{\modelname{} (Wan)} & $\times$ & \checkmark & \multirow{3}{*}{Pretrained Wan} & $[0]$ & 95.80 & 44.62 \\
 & $\times$ & \checkmark & & $[1,2]$ & 97.45 & 55.58 \\
 & $\times$ & \checkmark & & $[0,1,2]$ & 97.40 & 54.18 \\
\cmidrule{2-7}
 & $\times$ & \checkmark & Random Wan  & \multirow{2}{*}{$[1,2]$} & 97.65 & 58.63 \\
 & $\times$ & \checkmark & Pretrained VAE & & \textbf{98.15} & \textbf{64.54} \\
\midrule
\multirow{2}{*}{ImageWAM (Ovis-U1)} & $\times$ & $\times$ & -- & -- & 90.20 & 61.35 \\
 & \checkmark & $\times$ & -- & -- & 92.90 & 65.27 \\
\cmidrule{1-7}
\multirow{5}{*}{\modelname{} (Ovis-U1)} & $\times$ & \checkmark & \multirow{3}{*}{Pretrained Yak} & $[0]$ & 94.55 & 69.65 \\
 & $\times$ & \checkmark & & $[1]$ & \textbf{96.75} & \textbf{79.95} \\
 & $\times$ & \checkmark & & $[0,1]$ & 95.00 & 72.58 \\
\cmidrule{2-7}
 & $\times$ & \checkmark & Random Yak & \multirow{2}{*}{$[1]$} & 95.55 & 72.51 \\
 & $\times$ & \checkmark & Pretrained VAE & & 95.75 & 76.43 \\
\bottomrule
\end{tabular}
\vspace{-6mm}
\end{table}

Table~\ref{tab:objectives} compares generation with foresight and reports the readout offsets and target choices examined below. For FastWAM, replacing video denoising with foresight on pretrained Wan targets moves LIBERO from $97.00\%$ to $97.45\%$ and LIBERO-Plus from $55.31\%$ to $55.58\%$. The readout offset makes a larger difference than the objective swap itself: moving from offset $[0]$ to the future-only offsets $[1,2]$ raises LIBERO-Plus from $44.62\%$ to $55.58\%$, and adding the current position back, offsets $[0,1,2]$, lowers it to $54.18\%$. For the Wan anchor, the current-position feature $[0]$ may also contain future context through bidirectional attention.

For ImageWAM, the replacement improves LIBERO from $92.90\%$ to $96.75\%$ and LIBERO-Plus from $65.27\%$ to $79.95\%$. Future-offset targets again improve LIBERO-Plus success from $69.65\%$ at $[0]$ to $79.95\%$ at $[1]$. Figure~\ref{fig:pairedrollouts} shows two matched LIBERO-Long episodes. ImageWAM does not close the drawer in (a) and knocks over the second pot in (b), whereas \modelname{} completes both tasks. Appendix~\ref{app:morerollouts} gives more matched rollouts on LIBERO-Plus and RoboTwin 2.0.

The target rows of Table~\ref{tab:objectives} then vary the future anchor. On LIBERO-Plus, \modelname{} (Wan) performs best with the frozen VAE tokenizer alone, $64.54\%$, and a random Wan target still reaches $58.63\%$, whereas \modelname{} (Ovis-U1) performs best with the pretrained Yak anchor, $79.95\%$ against $76.43\%$ for the VAE tokenizer and $72.51\%$ for random Yak. The preferred target therefore differs between the two architectures. One possible explanation is the temporal information already present in the target representations. Their temporal inputs differ: Wan encodes a strided window over the 32-step action chunk, so its VAE latents already trace how the scene changes, whereas Yak encodes only the 16-step endpoint, which its pretrained layers relate to the current observation and instruction.

\begin{figure}[t]
\centering
\setlength{\tabcolsep}{1.6pt}
\renewcommand{\arraystretch}{1.05}
\begin{tabular}{@{}cccccc@{}}
\scriptsize Current frame
& \scriptsize Recorded future
& \scriptsize Recorded change
& \scriptsize ImageWAM
& \scriptsize \modelname{}, offset $[0]$
& \scriptsize \modelname{}, offset $[1]$ \\
\includegraphics[width=0.156\linewidth]{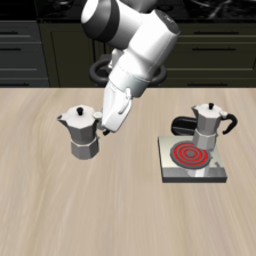}
& \includegraphics[width=0.156\linewidth]{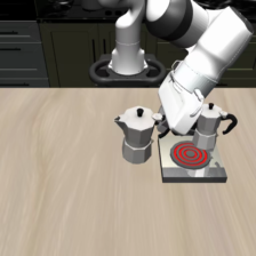}
& \includegraphics[width=0.156\linewidth]{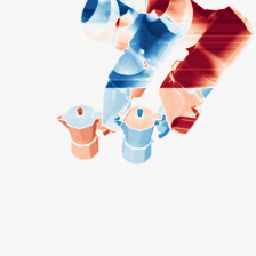}
& \includegraphics[width=0.156\linewidth]{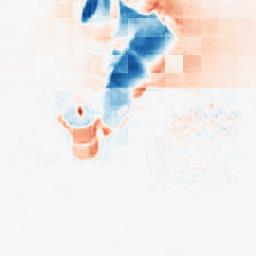}
& \includegraphics[width=0.156\linewidth]{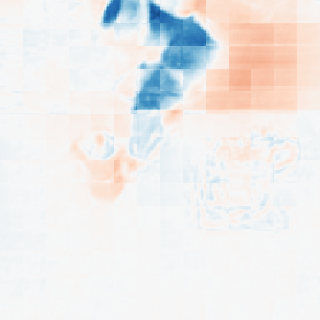}
& \includegraphics[width=0.156\linewidth]{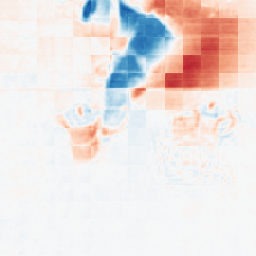} \\
\multicolumn{6}{c}{\scriptsize Darker \enspace \includegraphics[width=0.22\linewidth,height=0.55em]{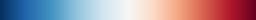} \enspace Brighter}
\end{tabular}
\vspace{-3mm}
\caption{Decoded future change for \emph{put both moka pots on the stove}, for ImageWAM and the two \modelname{} (Ovis-U1) policies of Table~\ref{tab:objectives} with readout offsets $[0]$ and $[1]$. Two MLP readouts reconstruct the current frame and the frame $16$ steps ahead from the same frozen features, and their difference is compared with the recorded change. Red marks pixels that get brighter and blue marks pixels that get darker, on a common scale. In this example, the features of \modelname{} trained with offset $[1]$ make the future-frame change more readily decodable than the features of ImageWAM or \modelname{} trained with offset $[0]$.}
\label{fig:readoutdiff}
\vspace{-5mm}
\end{figure}

What future changes can be decoded from these policy features? Two MLP readouts reconstruct the current frame and the frame $16$ steps ahead from each frozen policy. Figure~\ref{fig:readoutdiff} shows their signed difference for ImageWAM (Ovis-U1) and two \modelname{} variants. In this example, \modelname{} trained with offset $[1]$ recovers future changes more clearly. It also achieves the highest mean future-frame PSNR among the compared policies. These observations suggest that explicit future-position supervision helps make future information more accessible in current-observation features. Appendix~\ref{app:readoutdiff} describes the probes and reports the PSNR.

\subsection{Varying the backbone}
\label{sec:ports}

\begin{table}[t]
\newsavebox{\jamFactorialBox}
\newsavebox{\jamBackbonesBox}
\newlength{\jamFactorialRowGap}
\setlength{\jamFactorialRowGap}{0pt}
\newcommand{\jamFactorialTabular}{%
\begin{tabular}[b]{@{}llc@{}}
\toprule
Backbone & Anchor & LIBERO \\
\midrule
\noalign{\vskip 0.5\jamFactorialRowGap}
Pretrained & Pretrained & 97.45 \\
\addlinespace[\jamFactorialRowGap]
Pretrained & Random & \textbf{97.65} \\
\addlinespace[\jamFactorialRowGap]
Random & Random & 84.40 \\
\addlinespace[\jamFactorialRowGap]
Random & Pretrained & 71.15 \\
\noalign{\vskip 0.5\jamFactorialRowGap}
\bottomrule
\end{tabular}%
}
\sbox{\jamBackbonesBox}{%
\small
\setlength{\tabcolsep}{1.2pt}%
\begin{tabular}[b]{@{}lccc@{}}
\toprule
Policy backbone & Size & LIBERO & LIBERO-Plus \\
\midrule
ImageWAM (Ovis-U1)~\citep{imagewam2026} & 1.13B & 92.90 & 65.27 \\
\midrule
\modelname{} (Ovis-U1)~\citep{imagewam2026} & 1.13B & \textbf{96.75} & 79.95 \\
\modelname{} (DINOv2-B)~\citep{dinov2_2024} & 86.6M & 95.50 & 68.19 \\
\modelname{} (DINOv2-L)~\citep{dinov2_2024} & 304M & 96.30 & 70.39 \\
\modelname{} (Qwen3.5-2B)~\citep{qwen35_2026} & 2.0B & 96.55 & \textbf{81.14} \\
\bottomrule
\end{tabular}%
}
\sbox{\jamFactorialBox}{%
\small\setlength{\tabcolsep}{2.5pt}%
\jamFactorialTabular}
\setlength{\jamFactorialRowGap}{\dimexpr
  (\ht\jamBackbonesBox+\dp\jamBackbonesBox
   -\ht\jamFactorialBox-\dp\jamFactorialBox)/4\relax}
\sbox{\jamFactorialBox}{%
\small\setlength{\tabcolsep}{2.5pt}%
\jamFactorialTabular}
\noindent
\begin{minipage}[t]{0.28\textwidth}
\vspace{0pt}
\caption{\modelname{} (Wan) with the backbone and the future anchor each pretrained or random. }
\label{tab:factorial}
\end{minipage}\hfill%
\begin{minipage}[t]{0.68\textwidth}
\vspace{0pt}
\caption{\modelname{} across pretrained backbones with a common action-plus-foresight objective and frozen backbone copies as future anchors. Architecture and action-interface differences prevent isolating pretraining effects.}
\label{tab:imagewamarchmatched}
\end{minipage}
\par\nointerlineskip
\vspace{1.4mm}
\noindent
\makebox[0.28\textwidth][c]{\scalebox{0.95}{\usebox{\jamFactorialBox}}}\hfill%
\makebox[0.68\textwidth][c]{\scalebox{0.95}{\usebox{\jamBackbonesBox}}}
\vspace{-5mm}
\end{table}

Having replaced visual generation with foresight and compared different target representations, we next examine the role of backbone pretraining and whether the same recipe extends beyond generative backbones. Table~\ref{tab:factorial} separates backbone and future anchor initialization for \modelname{} (Wan). With a pretrained visual backbone, pretrained and random Wan targets reach $97.45\%$ and $97.65\%$ on LIBERO. With a random visual backbone, they reach $71.15\%$ and $84.40\%$. Pretrained backbone weights are important in both matched-target comparisons, whereas a random target remains effective. A frozen random future anchor supplies stable, input-dependent features of the recorded future, not independently sampled noise. Together with the VAE results in Table~\ref{tab:objectives}, this shows that the future anchor need not use the same pretrained backbone as the policy, while the policy backbone still relies on pretraining for strong control performance.

We next ask whether the recipe requires a generative backbone or can also be applied to other types of backbones. For vision-language Qwen3.5-2B, adding foresight raises LIBERO-Plus success from $77.03\%$ to $81.14\%$ with policy architecture and training settings fixed. Table~\ref{tab:imagewamarchmatched} compares the generative Ovis-U1, self-supervised DINOv2-B/L, and vision-language Qwen3.5-2B backbones under the same action-plus-foresight objective (Section~\ref{sec:instantiations}). \modelname{} (Ovis-U1) reaches $96.75\%$ on LIBERO and $79.95\%$ on LIBERO-Plus, and \modelname{} (Qwen3.5-2B) reaches $96.55\%$ and $81.14\%$. The two differ by $0.20$ and $1.19$ points despite their different pretraining objectives. \modelname{} (DINOv2-B) and \modelname{} (DINOv2-L) reach $68.19\%$ and $70.39\%$ on LIBERO-Plus. All four exceed the generative baseline ImageWAM (Ovis-U1) on both benchmarks. The recipe therefore extends beyond both visual generators and vision-language models, with the same action-only deployment path in every case. We also evaluate action-only training variants of Qwen3.5-2B and DINOv2-L in Appendix~\ref{app:actiononly}.

\begin{figure}[t]
\vspace{5.5mm}
\centering
\definecolor{wink}{HTML}{293340}\definecolor{wmuted}{HTML}{7A8590}\definecolor{wgray}{HTML}{7F8C96}\definecolor{worange}{HTML}{AD7158}
\definecolor{wblue}{HTML}{626B98}\definecolor{wgreen}{HTML}{427F7A}
\definecolor{worangel}{HTML}{D6B8AB}\definecolor{wbluel}{HTML}{B1B5CC}\definecolor{wgreenl}{HTML}{A1BFBD}
\begin{tikzpicture}[x=1cm,y=1cm]
  \tikzset{
    colhead/.style={font=\scriptsize\bfseries,text=wink,anchor=base,inner sep=0pt},
    rowlab/.style={font=\scriptsize,text=wink,rotate=90,anchor=center,inner sep=0pt},
    part/.style={font=\scriptsize\bfseries,text=wink,anchor=base west,inner sep=0pt},
    tiny/.style={font=\tiny,text=wink,inner sep=0pt},
  }
  \newcommand{\cell}[3]{\node[anchor=south west,inner sep=0pt] at (#1,#2) {\includegraphics[width=1.798cm]{Figures/attention/cells/#3}};}
  \node[part,anchor=base] at (4.666,4.23) {(a) Models with an Ovis-U1 backbone};
  \node[part,anchor=base] at (11.620,4.23) {(b) \modelname{} on other backbones};
  \foreach \x/\name in {0.42/Image,2.318/ImageWAM,4.716/ \modelname{},7.114/Difference,9.272/\modelname{} (DINOv2-L),11.670/\modelname{} (Qwen3.5-2B)}{
    \node[colhead] at ({\x+0.899},3.82) {\name};
  }
  \foreach \view/\Viewlab/\y in {agent/Agent/1.878,wrist/Wrist/0.02}{
    \node[rowlab] at (0.20,{\y+0.899}) {\Viewlab\ view};
    \cell{0.42}{\y}{M_\view_image.png}
    \cell{2.318}{\y}{M_\view_native.png}
    \cell{4.716}{\y}{M_\view_yak.png}
    \cell{7.114}{\y}{M_\view_diff.png}
    \cell{9.272}{\y}{M_\view_dinol.png}
    \cell{11.670}{\y}{M_\view_qwen.png}
  }
  \input{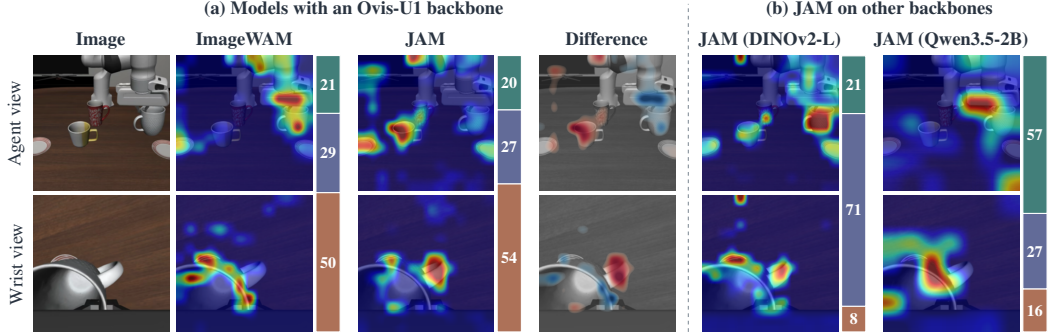}
  \draw[wmuted,line width=0.3pt,dash pattern=on 1.4pt off 1.4pt] (9.092,0.02) -- (9.092,4.36);
\end{tikzpicture}
\vspace{-3mm}
\caption{Action-to-visual attention on one LIBERO-Long observation for \emph{put the white mug on the left plate and put the yellow and white mug on the right plate}, in the agent and wrist views. (a) ImageWAM (Ovis-U1) and \modelname{} share the deployed policy architecture but use different training objectives. The difference map compares normalized display intensities, with red indicating higher values for \modelname{} and blue for ImageWAM. (b) \modelname{} on a self-supervised and a vision-language backbone. Bars show attention shares over text (orange), image (blue), and action keys (green), averaged over 100 reference frames.}
\label{fig:attention}
\vspace{-4mm}
\end{figure}

To see how these policies use their inputs, Figure~\ref{fig:attention} visualizes the action expert's attention to visual tokens on one observation across the different backbones. Each map averages heads, action queries, sampling steps, and layers, then shows the attention above that policy's mean map over $100$ reference frames, so it highlights content-dependent variation. On the displayed example, ImageWAM concentrates on the mug in the gripper, while \modelname{} on the same backbone places more attention on the second mug named in the instruction. The Difference panel makes this contrast clearer. The share bars summarize all $100$ frames and show that the action interfaces allocate attention differently: DINOv2-L gives its largest share to image keys, \modelname{} (Ovis-U1) to text keys, and Qwen3.5-2B to action keys. These maps and shares describe where attention goes. Appendix~\ref{app:attention} gives further details and results of this analysis.

\begin{table}[t]
\begin{minipage}[t]{0.48\textwidth}
\caption{RoboTwin 2.0 success (\%), averaged over Clean and Randomized settings (Table~\ref{tab:robotwinmain}).\strut}
\label{tab:mainrobotwin}
\centering
\small\setlength{\tabcolsep}{4pt}
\resizebox{\linewidth}{!}{\begin{tabular*}{213pt}{@{\extracolsep{\fill}}l c c@{}}
\toprule
Method & P.T. & AVG \\
\midrule
\multicolumn{3}{@{}l}{\emph{VLM-based policies}} \\
$\pi_0$ \citep{pi0_2024} & \checkmark & 62.2 \\
$\pi_{0.5}$ \citep{pi05_2025} & \checkmark & 79.8 \\
\modelname{} (Qwen3.5-2B)\smash{$^\dagger$} & $\times$ & \textbf{89.8}\\
\midrule
\multicolumn{3}{@{}l}{\emph{Video generator-based policies}} \\
Motus \citep{motus2026} & \checkmark & 87.8 \\
LingBot-VA \citep{lingbotva2026} & \checkmark & \textbf{92.2} \\
FastWAM\smash{$^\dagger$} \citep{fastwam2026} & $\times$ & 91.2 \\
\modelname{} (Wan)\smash{$^\dagger$} & $\times$ & 91.4 \\
\midrule
\multicolumn{3}{@{}l}{\emph{Image generator-based policies}} \\
ImageWAM (Ovis-U1)\smash{$^\dagger$} \citep{imagewam2026} & $\times$ & 89.0 \\
ImageWAM (FLUX.2)\smash{$^\dagger$} \citep{imagewam2026} & $\times$ & 92.9 \\
\modelname{} (Ovis-U1)\smash{$^\dagger$} & $\times$ & 89.9 \\
\modelname{} (FLUX.2)\smash{$^\dagger$} & $\times$ & \textbf{93.1} \\
\bottomrule
\end{tabular*}}
\end{minipage}\hfill
\begin{minipage}[t]{0.48\textwidth}
\caption{LIBERO success (\%), averaged over four suites. Table~\ref{tab:libero} gives per-suite results.\strut}
\label{tab:mainlibero}
\centering
\small\setlength{\tabcolsep}{4pt}
\resizebox{\linewidth}{!}{\begin{tabular*}{213pt}{@{\extracolsep{\fill}}l c c@{}}
\toprule
Method & P.T. & AVG \\
\midrule
\multicolumn{3}{@{}l}{\emph{VLM-based policies}} \\
OpenVLA \citep{openvla2024} & \checkmark & 76.5 \\
GR00T N1 \citep{gr00tn1_2025} & \checkmark & 93.9 \\
$\pi_0$ \citep{pi0_2024} & \checkmark & 94.1 \\
$\pi_{0.5}$ \citep{pi05_2025} & \checkmark & \textbf{96.9} \\
\modelname{} (Qwen3.5-2B)\smash{$^\dagger$} & $\times$ & 96.6 \\
\midrule
\multicolumn{3}{@{}l}{\emph{Video generator-based policies}} \\
Motus \citep{motus2026} & \checkmark & 97.7 \\
LingBot-VA \citep{lingbotva2026} & \checkmark & \textbf{98.5} \\
FastWAM\smash{$^\dagger$} \citep{fastwam2026} & $\times$ & 97.0 \\
\modelname{} (Wan)\smash{$^\dagger$} & $\times$ & 98.2 \\
\midrule
\multicolumn{3}{@{}l}{\emph{Image generator-based policies}} \\
ImageWAM (Ovis-U1)\smash{$^\dagger$} \citep{imagewam2026} & $\times$ & 92.9 \\
\modelname{} (Ovis-U1)\smash{$^\dagger$} & $\times$ & \textbf{96.8} \\
\bottomrule
\end{tabular*}}
\end{minipage}%
\vspace{-4.5mm}
\end{table}

\subsection{Benchmark comparisons}
\label{sec:published}

Tables~\ref{tab:mainrobotwin}, \ref{tab:mainlibero}, and~\ref{tab:mainplus} compare policies on RoboTwin 2.0, LIBERO, and LIBERO-Plus, grouped by backbone pretraining family, with bold marking the best score within each group. The P.T.\ column marks embodied pretraining on robot data, which none of our policies use. Rows marked $\dagger$ are evaluated by us, and the others keep their published values and protocols. \modelname{} (Wan) and \modelname{} (Ovis-U1) use the configurations selected in Table~\ref{tab:objectives}. Full results appear in Appendices~\ref{app:liberoplus} and~\ref{app:robotwin}.

On RoboTwin 2.0, \modelname{} has the best score among VLM-based and image generator-based policies, and \modelname{} (FLUX.2) has the highest listed score, $93.1\%$. Among video generator-based policies, LingBot-VA, which uses embodied pretraining, leads \modelname{} (Wan) by $0.8$ points. Unlike LingBot-VA, \modelname{} (Wan) predicts actions without first generating future video, and \citet{fastwam2026} measured this inference path as over four times faster than a matched variant that generates video first. On LIBERO, \modelname{} (Ovis-U1) leads the image generator-based policies, while $\pi_{0.5}$ and LingBot-VA, both with embodied pretraining, stay $0.3$ points ahead of \modelname{} (Qwen3.5-2B) and \modelname{} (Wan). On LIBERO-Plus, \modelname{} has the best average in all three groups. It exceeds FastWAM and ImageWAM (Ovis-U1) by $9.2$ and $14.7$ points and trails them only in the background and robot-state categories, respectively. Overall, \modelname{} matches or exceeds its generative baselines on all three benchmarks, so learning from future observations does not require a visual generative objective. \modelname{} (Qwen3.5-2B) is competitive on LIBERO and leads on LIBERO-Plus, but trails \modelname{} (FLUX.2) by $3.30$ percentage points on RoboTwin 2.0.
These results support a common recipe for vision-language and generative backbones: action and foresight losses during training, with only the action policy retained at inference.

\begin{table}[t]
\caption{LIBERO-Plus success (\%) by perturbation category, with the task-weighted average. Published rows retain their original evaluation protocols; † marks our evaluation on a fixed 1,506-task subset.}
\vspace{2mm}
\label{tab:mainplus}
\centering
\small\setlength{\tabcolsep}{3pt}
\resizebox{\textwidth}{!}{%
\begin{tabular}{@{}l c ccccccc c@{}}
\toprule
Method & P.T. & Camera & Robot & Language & Light & Background & Noise & Layout & AVG \\
\midrule
\multicolumn{10}{@{}l}{\emph{VLM-based policies}} \\
UniVLA \citep{univla2025} & \checkmark & 1.8 & 46.2 & 69.6 & 69.0 & 81.0 & 21.2 & 31.9 & 42.9 \\
OpenVLA-OFT \citep{openvlaoft2025} & \checkmark & 56.4 & 31.9 & \textbf{79.5} & 88.7 & \textbf{93.3} & 75.8 & 74.2 & 69.6 \\
$\pi_0$ \citep{pi0_2024} & \checkmark & 13.8 & 6.0 & 58.8 & 85.0 & 81.4 & 79.0 & 68.9 & 53.6 \\
$\pi_0$-Fast \citep{pi0fast2025} & \checkmark & 65.1 & 21.6 & 61.0 & 73.2 & 73.2 & 74.4 & 68.8 & 61.6 \\
WorldVLA \citep{worldvla2025} & $\times$ & 0.1 & 27.9 & 41.6 & 43.7 & 17.1 & 10.9 & 38.0 & 25.0 \\
\modelname{} (Qwen3.5-2B)$^\dagger$ & $\times$ & \textbf{80.8} & \textbf{61.4} & 71.7 & \textbf{93.8} & 81.7 & \textbf{94.2} & \textbf{84.5} & \textbf{81.1} \\
\midrule
\multicolumn{10}{@{}l}{\emph{Video generator-based policies}} \\
FastWAM$^\dagger$ \citep{fastwam2026} & $\times$ & 24.2 & 50.3 & 78.7 & 87.6 & \textbf{49.5} & 39.9 & 67.3 & 55.3 \\
\modelname{} (Wan)$^\dagger$ & $\times$ & \textbf{25.4} & \textbf{68.0} & \textbf{92.6} & \textbf{93.2} & 46.2 & \textbf{52.7} & \textbf{81.4} & \textbf{64.5} \\
\midrule
\multicolumn{10}{@{}l}{\emph{Image generator-based policies}} \\
ImageWAM (Ovis-U1)$^\dagger$ \citep{imagewam2026} & $\times$ & 57.9 & \textbf{54.8} & 67.6 & 78.9 & 66.7 & 68.6 & 65.0 & 65.3 \\
\modelname{} (Ovis-U1)$^\dagger$ & $\times$ & \textbf{83.8} & 49.2 & \textbf{84.8} & \textbf{95.7} & \textbf{80.1} & \textbf{91.9} & \textbf{72.3} & \textbf{79.9} \\

\bottomrule
\end{tabular}}
\vspace{-2mm}
\end{table}

\subsection{Training cost}
\label{sec:cost}
\modelname{} reduces online training cost by processing only the current observation in the trainable visual branch and caching fixed targets. With cached foresight targets, the estimated cost per sample per update falls from $15.52$ TFLOPs for FastWAM to $4.19$ for \modelname{} (Wan), a $73.0\%$ reduction. Generation denoises a nine-frame window that the video VAE must also encode, whereas cached foresight uses only the current frame online. With cached Yak targets, the estimated cost falls from $5.81$ TFLOPs for ImageWAM (Ovis-U1) to $3.67$ for \modelname{} (Ovis-U1), a $36.8\%$ reduction. Its generative branch predicts only one future frame, and both objectives process the text context. Measured on one GH200, the GPU time of the operations covered by these estimates falls by $67.3\%$ and $37.8\%$ in the same two comparisons. Appendix~\ref{app:costacct} details the FLOP estimates and GPU-time measurements.

\section{Conclusion}

\modelname{} bridges vision-language-action and world-action models by separating backbone initialization, supervision on future observations, and the modules retained at inference. It trains a pretrained visual backbone and an action expert with action and foresight losses, using frozen future representations as targets and discarding the foresight branch after training. With the policy architecture fixed, replacing visual generation with foresight yields higher success rates than the corresponding generative baselines on LIBERO and LIBERO-Plus, with the largest gains on LIBERO-Plus; RoboTwin 2.0 success is similar or slightly higher. With cached targets, \modelname{} reduces online FLOPs per sample per update by $73.0\%$ relative to FastWAM and $36.8\%$ relative to ImageWAM. With the recipe fixed, the same supervision applies to generative, self-supervised, and vision-language backbones, and the best generator-initialized and vision-language policies achieve comparable performance. These results show that effective foresight training is possible with either pretraining family, while the cross-backbone comparison does not isolate the effect of pretraining alone. Both families can be trained as \modelfullname{}s that learn from recorded futures without visual generation during policy training or additional inference machinery.

\clearpage
\subsection*{AI use statement}
We used generative AI tools to assist with code implementation and literature checks, obtain reviewer-style research feedback, and refine the writing. The authors take full responsibility for the final text, code, figures, methods, results, analyses, claims, and references, including material prepared with AI assistance.

\subsection*{Ethics statement}
This study evaluates robot manipulation policies in simulation using public models, code, and datasets. Experiments use LIBERO, LIBERO-Plus, and RoboTwin 2.0 and involve no human participants or personal data. Relevant risks include failures under distribution shift and unsafe deployment of simulation-trained policies without physical validation.

\subsection*{Reproducibility statement}
Section~\ref{sec:method} defines \modelname{}'s objective and policy interfaces, and Section~\ref{sec:setup} describes the benchmarks and comparison protocol. Appendix~\ref{app:impl} specifies the architectures, target construction, training hyperparameters, and budgets. Appendix~\ref{app:costacct} details the FLOP estimates and GPU-time measurements, while Appendix~\ref{app:config} reports the ablations and action-only controls. Appendix~\ref{app:inprogress} documents the evaluation and analysis protocols, including LIBERO-Plus subset sampling, paired statistical tests, and independent training runs. Our implementation builds on public FastWAM and ImageWAM code and public benchmark datasets. Our code and selected model weights are publicly available.


\bibliography{reference}
\bibliographystyle{iclr2027_conference}


\clearpage
\appendix
\section{Implementation details}
\label{app:impl}

\subsection{Training setup}
\label{app:training}

All runs use eight NVIDIA GH200 GPUs and bf16 mixed precision. We use AdamW with betas $(0.9,0.95)$ and weight decay $0.01$, a learning rate of $10^{-4}$ with cosine decay and $5\%$ warmup, and gradient clipping at $1.0$.

With Wan, \modelname{} follows the FastWAM recipe: ZeRO-1, 32-step action chunks, and targets from a recorded future window. Ovis-U1, FLUX.2, Qwen3.5-2B, and DINOv2-B/L follow the ImageWAM recipe: ZeRO-0, 16-step action chunks, and targets from the recorded endpoint. Future observations enter only the future-anchor branch.

\paragraph{LIBERO and LIBERO-Plus.}
Both recipes use two $224{\times}224$ camera views and global batch size 128. FastWAM trains for $20$k updates and ImageWAM for $21.7$k.

\paragraph{RoboTwin 2.0.}
All policies use the same three camera views. FastWAM combines them into a $384{\times}320$ image, and Ovis-U1 and FLUX.2 into a $288{\times}256$ image. Qwen3.5-2B encodes the three views separately at $224{\times}224$ per view. All policies are trained for five epochs. FastWAM uses global batch size $1{,}024$ and $29{,}355$ updates; the ImageWAM recipe uses global batch size 384 and $69{,}900$ updates.

\subsection{Backbones and target interfaces}
\label{app:interfaces}

Table~\ref{tab:targetinterfaces} summarizes the training targets and action contexts for each backbone.

\paragraph{Wan.}
FastWAM pairs a 30-block Wan2.2-5B backbone with a matching action expert through directional mixed attention \citep{wan2025,fastwam2026}. The two LIBERO views form a $224{\times}448$ frame with 98 current-frame latent tokens. The nine-frame target window becomes three temporal VAE positions; readouts $[1,2]$ span steps 4--16 and 20--32. Alignment uses fixed spatial indices without object tracking. VAE-direct targets lift each $2{\times}2{\times}48$ patch from 192 to 3,072 dimensions by fixed coordinate repetition and scaling, preserving norms and cosine similarities. The frozen Wan future anchors in Table~\ref{tab:objectives} use pretrained weights or default random initialization. The EMA future anchor in Table~\ref{tab:teacherfull} uses an FP32 copy of the policy backbone updated with decay $0.995$.
For the readout ablation, the pretrained Wan future anchor receives the same full clean window at diffusion time $t=0$. We align policy predictions to its current-position features at $[0]$, future-position features at $[1,2]$, or both at $[0,1,2]$. Bidirectional attention can encode future context implicitly at $[0]$. Future-only supervision explicitly targets the future positions at $[1,2]$.

\paragraph{Ovis-U1.}
We retain ImageWAM's Yak18 backbone, with six double-stream and twelve single-stream blocks at width 1,536, and its matching action expert \citep{imagewam2026}. The expert attends to text, condition-image, and action keys. Two LIBERO views form a $224{\times}448$ frame with $14{\times}28$ packed Ovis-VAE tokens; a frozen Ovis stack supplies text features. The frozen Yak future anchor reads the current frame, text, and 16-step future frame; target tokens are excluded from the trainable stream. Random-Yak and VAE-direct controls keep readouts and training budgets fixed. VAE-direct lifts 16-dimensional tokens to 1,536 dimensions using Wan's norm-preserving map, without a target transformer or text conditioning.

\paragraph{FLUX.2.}
We use ImageWAM's FLUX.2 Klein 4B backbone with five double-stream and twenty single-stream blocks, paired with a matching action expert of width 1,024. Instructions are 512 cached tokens from a frozen Qwen3-4B text encoder. The frozen FLUX.2 future anchor reads the current frame, instruction features, and 16-step future frame at diffusion time $t=0$.

\paragraph{Qwen3.5-2B.}
The 24-layer backbone encodes each view separately alongside the instruction, without a separate text encoder. Its action expert copies compatible weights from 18 GatedDeltaNet and six full-attention blocks and adds action input, output, and flow-time modules. Causal token mixers process the current image--text prefix before action tokens, preventing actions from influencing the prefix. The frozen future anchor reads the instruction, current views, and 16-step future views in one causal sequence. Future-image features supervise current-image features at the same layer; future views never enter the policy prefix.

\paragraph{DINOv2-B/L.}
Each encoder processes views separately. Its action expert matches its depth and width, copies normalization, attention, and MLP weights, and adds action input, position, flow-time, AdaLN, and output modules. Action queries attend to the text prefix, same-layer patch keys and values from both views, and action tokens; visual queries never attend to text or actions. The text prefix contains 512 cached tokens from frozen Qwen3-4B, projected to the action width. A frozen DINOv2 copy encodes each future view into final-layer patch targets for the corresponding policy layer.

\begin{table}[htbp]
\vspace{-2mm}
\caption{Training targets and action contexts across backbones. Frame offsets are measured in environment steps relative to the current observation; Layer reports the supervised policy block / total depth.}
\vspace{2mm}
\label{tab:targetinterfaces}
\centering
\small
\setlength{\tabcolsep}{4pt}
\renewcommand{\arraystretch}{1.2}
\begin{tabular*}{\textwidth}{@{\extracolsep{\fill}}lcccc@{}}
\toprule
\multirow{2}{*}{Future anchor} & \multicolumn{3}{c}{\textit{Training targets}} & \multicolumn{1}{c}{\textit{Action inference}} \\
\cmidrule(lr){2-4}\cmidrule(l){5-5}
 & Frames & Layer & Readout & Context \\
\midrule
Wan & \multirow{2}{*}{$0,4,\ldots,32$} & \multirow{2}{*}{$18/30$} & \multirow{2}{*}{$[1,2]$} & \multirow{2}{*}{Wan + UMT5} \\
Wan VAE & & & & \\
\midrule
Yak18 & \multirow{2}{*}{$0,16$} & $11/18$ & \multirow{2}{*}{$[1]$} & Yak18 + Ovis \\
FLUX.2 & & $15/25$ & & FLUX.2 + Qwen3-4B \\
\midrule
Qwen3.5-2B & $0,16$ & $13/24$ & $[1]$ & Causal image--text prefix \\
\midrule
DINOv2-B & \multirow{2}{*}{$16$} & $12/12$ & \multirow{2}{*}{$[1]$} & \multirow{2}{*}{DINOv2 + Qwen3-4B} \\
DINOv2-L & & $24/24$ & & \\
\bottomrule
\end{tabular*}
\vspace{-2mm}
\end{table}

\FloatBarrier

\subsection{Prediction head and loss}
\label{app:predictionhead}

Each head uses a learned $D\!\to\!d$ input projection, two shared pre-LayerNorm transformer blocks, and one linear $d\!\to\!D_g$ readout per target offset. Wan also normalizes the trunk output. Token count and spatial order are preserved. Table~\ref{tab:headdims} gives the dimensions; $D_g$ is measured after fixed lifting for VAE-direct targets. Offset ablations keep the trunk fixed and use $|\mathcal K|$ readouts.

Equation~\ref{eq:total} combines the action and foresight losses. The foresight term in Equation~\ref{eq:fra} averages cosine distance over valid examples, tokens, and readout offsets. Except in the loss-weight ablation, its total weight is $\lambda_{\mathrm F}=0.6$ for FastWAM and $0.3$ for other policies, independent of the number of offsets. Targets receive no gradients; the anchor branch and prediction head are removed after training.

\begin{table}[htbp]
\vspace{-2mm}
\caption{Default prediction-head dimensions. Layers, Heads, and FFN describe the shared transformer trunk; Readouts denotes the number of supervised offsets.}

\label{tab:headdims}
\begin{center}
\small\setlength{\tabcolsep}{5pt}
\begin{tabular}{@{}lrrrrrrr@{}}
\toprule
Policy & Input $D$ & Target $D_g$ & Trunk $d$ & Layers & Heads & FFN & Readouts \\
\midrule
\modelname{} (Wan) & 3072 & 3072 & 1536 & 2 & 24 & 6144 & 2 \\
\modelname{} (Ovis-U1) & 1536 & 1536 & 768 & 2 & 12 & 3072 & 1 \\
\modelname{} (FLUX.2) & 3072 & 3072 & 1536 & 2 & 24 & 6144 & 1 \\
\modelname{} (Qwen3.5-2B) & 2048 & 2048 & 1024 & 2 & 16 & 4096 & 1 \\
\modelname{} (DINOv2-B) & 768 & 768 & 384 & 2 & 6 & 1536 & 1 \\
\modelname{} (DINOv2-L) & 1024 & 1024 & 512 & 2 & 8 & 2048 & 1 \\
\bottomrule
\end{tabular}
\end{center}
\vspace{-2mm}
\end{table}

\section{Training-cost accounting}
\label{app:costacct}

\subsection{Parameters}

Table~\ref{tab:costbreakdown} counts training-only parameters for Wan with VAE-direct $[1,2]$ targets and Ovis-U1 with pretrained Yak $[1]$ targets. Deployed parameters match the respective generative baselines. Wan reuses the policy's frozen VAE and needs no additional target parameters; Ovis-U1 loads a frozen $1.13$B Yak18 only for online targets. Head dimensions are given in Table~\ref{tab:headdims}.

\subsection{FLOPs}

For Table~\ref{tab:costbreakdown}, we estimate a forward pass as $2NT$ FLOPs for $N$ active parameters and $T$ tokens. Trainable blocks cost three forward-equivalents, or four with ImageWAM's activation checkpointing; frozen anchor backbones and VAEs run forward only. We omit attention-score terms, text encoding, and optimizer updates.

Wan's $4.9$B visual expert processes 294 tokens under generation and 98 under foresight, plus key and value projections of 129 context tokens per block. Online VAE-direct targets require encoding the full window; caching requires only the current frame. For Ovis-U1, image and action tokens traverse $566$M parameters and text tokens $573$M. Observation and generation-target sequences each contain 392 tokens; text context averages 296 tokens. Online targets run frozen Yak18 to block 11. VAE FLOPs and Wan action-expert and head FLOPs are measured at the operator level.

We profile GPU time with the CUDA activity profiler on one GH200 at batch 16, using activation checkpointing only for Ovis-U1 under both objectives. Wan timings cover forward and backward passes. Ovis-U1 timings subtract $0.251$ s for frozen prompt encoding and $0.109$ s for optimizer updates, both unchanged between objectives. Wan's measured reductions fall below the FLOP estimates because normalization, activation, and rotary-embedding kernels shrink less than token counts. Cached costs exclude one-time target extraction; cache reuse requires fixed future anchors, preprocessing, temporal indices, and conditioning.

\begin{table}[htbp]
\vspace{-2mm}
\caption{Training-only parameters, estimated online FLOPs per sample per update, and measured GPU time over the profiled operations.}
\vspace{2mm}
\label{tab:costbreakdown}
\centering
\small\setlength{\tabcolsep}{4pt}
\resizebox{\textwidth}{!}{%
\begin{tabular}{@{}llcccccccc c cc@{}}
\toprule
 & & \multicolumn{2}{c}{Training Params. (M)} & \multicolumn{7}{c}{TFLOPs per sample per update} & \multicolumn{2}{c}{Measured GPU time} \\
\cmidrule(lr){3-4}\cmidrule(lr){5-11}\cmidrule(lr){12-13}
Policy & Targets & Head & Anchor & Visual & Action & Head & Target & VAE & Total & Reduction & s / step & Reduction \\
\midrule
FastWAM & -- & -- & -- & $8.93$ & $0.05$ & -- & -- & $6.54$ & $15.52$ & -- & $0.789$ & -- \\
\modelname{} (Wan) & online & $71$ & $0$ & $3.17$ & $0.05$ & $0.03$ & $0$ & $6.54$ & $9.80$ & $36.9\%$ & $0.532$ & $32.6\%$ \\
\modelname{} (Wan) & cached & $71$ & $0$ & $3.17$ & $0.05$ & $0.03$ & $0$ & $0.93$ & $4.19$ & $73.0\%$ & $0.258$ & $67.3\%$ \\
\midrule
ImageWAM (Ovis-U1) & -- & -- & -- & $4.91$ & $0.07$ & -- & -- & $0.83$ & $5.81$ & -- & $0.473$ & -- \\
\modelname{} (Ovis-U1) & online & $16.5$ & $1{,}127$ & $3.13$ & $0.07$ & $0.05$ & $0.80$ & $0.83$ & $4.88$ & $15.9\%$ & $0.403$ & $14.8\%$ \\
\modelname{} (Ovis-U1) & cached & $16.5$ & $0$ & $3.13$ & $0.07$ & $0.05$ & $0$ & $0.41$ & $3.67$ & $36.8\%$ & $0.294$ & $37.8\%$ \\
\bottomrule
\end{tabular}}
\vspace{-2mm}
\end{table}

\section{Ablations for foresight supervision}
\label{app:config}

Unless varied below, Wan foresight ablations use frozen pretrained targets, block 18, equally weighted readout offsets $[1,2]$, $\lambda_{\mathrm F}=0.6$, a width-1,536 transformer head, and $20$k updates. Success rates are reported in percent.

\subsection{EMA future anchor}
\label{app:variants}

We compare the frozen pretrained Wan future anchor with an exponential moving average of the policy backbone to test whether updating the target during training improves control. In Table~\ref{tab:teacherfull}, the EMA target reaches $96.40\%$ on LIBERO, below the frozen pretrained target's $97.45\%$.

\begin{table}[htbp]
\vspace{-2mm}
\caption{Frozen pretrained and EMA Wan future anchors on LIBERO. Success rates are in percent.}
\vspace{2mm}
\label{tab:teacherfull}
\centering
\small\setlength{\tabcolsep}{4pt}
\begin{tabular}{lc ccccc}
\toprule
Policy & Future anchor & Spatial & Object & Goal & Long & AVG \\
\midrule
FastWAM & -- & 97.4 & 99.6 & 96.6 & 94.4 & 97.00 \\
FastWAM action-only & -- & 95.0 & 99.6 & 94.4 & 92.0 & 95.25 \\
\midrule
\multirow{2}{*}{\modelname{} (Wan)} & Pretrained Wan & \textbf{98.2} & \textbf{100.0} & \textbf{97.6} & 94.0 & \textbf{97.45} \\
& Online Wan EMA & 95.0 & 99.6 & 96.4 & \textbf{94.6} & 96.40 \\
\bottomrule
\end{tabular}
\vspace{-2mm}
\end{table}

\subsection{Supervised depth}
We apply foresight supervision at different Wan blocks to examine how the supervised depth affects control. In Table~\ref{tab:depth}, blocks 14, 16, and 18 reach $97.10\%$--$97.45\%$ on LIBERO, with block 18 performing best. 

\begin{table}[ht]
\vspace{-2mm}
\caption{LIBERO success in percent by Wan block. Depth indexes the 30 visual blocks.}
\label{tab:depth}
\begin{center}
\small\setlength{\tabcolsep}{4pt}
\begin{tabular}{lc ccccc}
\toprule
Policy & Depth & Spatial & Object & Goal & Long & AVG \\
\midrule
FastWAM & -- & 97.4 & 99.6 & 96.6 & \textbf{94.4} & 97.00 \\
FastWAM action-only & -- & 95.0 & 99.6 & 94.4 & 92.0 & 95.25 \\
\midrule
\multirow{4}{*}{\modelname{} (Wan)} & $14$ & 98.0 & 99.8 & 96.4 & 94.2 & 97.10 \\
& $16$ & \textbf{99.2} & 99.8 & 96.2 & 94.0 & 97.30 \\
& $\mathbf{18}$ & 98.2 & \textbf{100.0} & \textbf{97.6} & 94.0 & \textbf{97.45} \\
& $20$ & 96.8 & 99.2 & 95.4 & 90.4 & 95.45 \\
\bottomrule
\end{tabular}
\end{center}
\vspace{-2mm}
\end{table}

\subsection{Loss weight}
We vary $\lambda_{\mathrm F}$ to change the strength of foresight supervision relative to the action loss. The sweep in Table~\ref{tab:lambdaweight} gives the highest average at $\lambda_{\mathrm F}=0.6$, the only tested setting to exceed FastWAM. Long is most sensitive, ranging from $90.0\%$ to $97.0\%$. As specified in Appendix~\ref{app:predictionhead}, each readout offset has weight $\lambda_{\mathrm F}/|\mathcal K|$, so adding offsets does not increase the total weight.

\begin{table}[ht]
\vspace{-2mm}
\caption{LIBERO success rates in percent across foresight loss weights $\lambda_{\mathrm F}$ for Wan.}
\label{tab:lambdaweight}
\begin{center}
\small\setlength{\tabcolsep}{5pt}
\begin{tabular}{lc ccccc}
\toprule
Policy & $\lambda_{\mathrm F}$ & Spatial & Object & Goal & Long & AVG \\
\midrule
FastWAM & -- & 97.4 & 99.6 & 96.6 & 94.4 & 97.00 \\
FastWAM action-only & -- & 95.0 & 99.6 & 94.4 & 92.0 & 95.25 \\
\midrule
\multirow{4}{*}{\modelname{} (Wan)} & $0.4$ & 96.8 & 99.4 & 96.6 & 90.0 & 95.70 \\
& $0.5$ & 95.6 & 99.8 & 95.2 & \textbf{97.0} & 96.90 \\
& $\mathbf{0.6}$ & \textbf{98.2} & \textbf{100.0} & \textbf{97.6} & 94.0 & \textbf{97.45} \\
& $0.7$ & 97.4 & 99.8 & 96.0 & 93.8 & 96.75 \\
\bottomrule
\end{tabular}
\end{center}
\vspace{-2mm}
\end{table}

\FloatBarrier

\subsection{Prediction-head capacity}
We compare transformer and MLP prediction heads at different widths to assess the effect of head architecture and capacity. We use the 71M transformer head with width 1,536, which reaches the highest success rate of $97.45\%$ in Table~\ref{tab:head}. Larger heads do not consistently improve success. Transformers exceed MLPs by $2.40$ and $0.95$ points at widths 1,536 and 3,072, respectively.

\begin{table}[ht]
\vspace{-2mm}
\caption{Prediction-head capacity on LIBERO. Heads use 3,072-D inputs and outputs, two-layer trunks, and two readouts. Transformers have 24 attention heads. Their input projections are learned except at width 3,072, which uses the identity.}
\vspace{2mm}
\label{tab:head}
\centering
\small\setlength{\tabcolsep}{3pt}
\resizebox{\textwidth}{!}{\begin{tabular}{lc c c cccccc}
\toprule
Policy & Head & Width & FFN & Params (M) & Spatial & Object & Goal & Long & AVG \\
\midrule
FastWAM & -- & -- & -- & -- & 97.4 & 99.6 & 96.6 & \textbf{94.4} & 97.00 \\
FastWAM action-only & -- & -- & -- & -- & 95.0 & 99.6 & 94.4 & 92.0 & 95.25 \\
\midrule
\multirow{5}{*}{\modelname{} (Wan)} & \multirow{3}{*}{Transformer} & 3072 & 4096 & 145 & 97.4 & \textbf{100.0} & 96.2 & 91.6 & 96.30 \\
& & 1536 & 6144 & 71 & \textbf{98.2} & \textbf{100.0} & \textbf{97.6} & 94.0 & \textbf{97.45} \\
& & 768 & 6144 & 31 & 96.4 & \textbf{100.0} & 97.0 & 89.6 & 95.75 \\
\cmidrule{2-10}
 & \multirow{2}{*}{MLP} & 3072 & -- & 38 & 95.8 & \textbf{100.0} & 94.2 & 91.4 & 95.35 \\
 & & 1536 & -- & 17 & 95.8 & 99.2 & 94.0 & 91.2 & 95.05 \\
\bottomrule
\end{tabular}}
\vspace{-2mm}
\end{table}

\FloatBarrier

\subsection{Action-only controls}
\label{app:actiononly}

To measure the contribution of foresight, we compare training with and without it for Qwen3.5-2B and DINOv2-L. Each pair in Table~\ref{tab:qwenactiononly} uses the same architecture and training settings. For Qwen3.5-2B, foresight raises LIBERO-Plus success from $77.03\%$ to $81.14\%$, with the largest gains in background and layout; robot-state and language perturbations favor action-only training. For DINOv2-L, foresight raises success from $68.26\%$ to $70.39\%$, mainly through camera and robot-state gains. Language, light, and background favor action-only training.

\begin{table}[htbp]
\vspace{-2mm}
\caption{Action-only controls on LIBERO-Plus. Success rates are in percent; AVG is micro-averaged. Bold marks the higher score within each backbone pair.}
\vspace{2mm}
\label{tab:qwenactiononly}
\centering
\small
\setlength{\tabcolsep}{4pt}
\renewcommand{\arraystretch}{1.15}
\begin{tabular*}{\textwidth}{@{\extracolsep{\fill}}l ccccccc c@{}}
\toprule
Method & Camera & Robot & Language & Light & Background & Noise & Layout & AVG \\
\midrule


Qwen3.5-2B action-only & 80.00 & \textbf{68.02} & \textbf{72.54} & 92.55 & 54.84 & 93.02 & 75.45 & 77.03 \\
\modelname{} (Qwen3.5-2B) & \textbf{80.83} & 61.42 & 71.72 & \textbf{93.79} & \textbf{81.72} & \textbf{94.19} & \textbf{84.55} & \textbf{81.14} \\
\midrule
DINOv2-L action-only & 37.92 & 32.99 & \textbf{86.48} & \textbf{86.96} & \textbf{85.48} & 78.29 & 72.73 & 68.26 \\
\modelname{} (DINOv2-L) & \textbf{55.42} & \textbf{39.09} & 80.74 & 85.71 & 77.42 & \textbf{81.40} & \textbf{73.18} & \textbf{70.39} \\
\bottomrule
\end{tabular*}
\vspace{-2mm}
\end{table}

\subsection{Separate action expert or shared weights}
\label{app:sharedweights}

We test whether foresight training requires a separate action expert by sharing the Qwen3.5 backbone weights for action prediction. The 3.60B policy includes a 2.21B backbone, a 1.38B action expert, and a small state projector. The expert copies decoder blocks without duplicating the vision encoder or token embeddings. With the objective and training budget fixed, Table~\ref{tab:sharedweights} reports $81.14\%$ LIBERO-Plus success with the expert, versus $79.61\%$ and $80.68\%$ with shared 2B and 4B backbones. The recipe also works without a separate expert. In these single-run comparisons, the expert achieves similar observed success to shared 4B with fewer parameters.

\begin{table}[htbp]
\vspace{-2mm}
\caption{Separate action expert versus shared backbone weights for Qwen3.5 on LIBERO-Plus. Success rates are in percent; AVG is task-weighted and Params counts deployed parameters.}
\vspace{2mm}
\label{tab:sharedweights}
\centering
\small
\setlength{\tabcolsep}{2.5pt}
\begin{tabular*}{\textwidth}{@{\extracolsep{\fill}}l l c ccccccc c@{}}
\toprule
Policy & Actions & Params & Camera & Robot & Language & Light & Background & Noise & Layout & AVG \\
\midrule
\modelname{} (Qwen3.5-2B) & Expert & 3.60B & \textbf{80.8} & 61.4 & 71.7 & \textbf{93.8} & \textbf{81.7} & 94.2 & \textbf{84.5} & \textbf{81.14} \\
\midrule
\modelname{} (Qwen3.5-2B) & Shared & 2.22B & 80.4 & 64.0 & 68.9 & 92.5 & 76.9 & \textbf{96.1} & 78.2 & 79.61 \\
\modelname{} (Qwen3.5-4B) & Shared & 4.55B & 80.4 & \textbf{67.5} & \textbf{74.6} & 91.9 & 76.3 & 95.3 & 77.7 & 80.68 \\
\bottomrule
\end{tabular*}
\vspace{-2mm}
\end{table}

\subsection{Training budget}
To assess sensitivity to training duration, we train FastWAM and \modelname{} with Wan for $10$k and $20$k updates. In Table~\ref{tab:budget}, \modelname{} with Wan exceeds $20$k FastWAM on LIBERO after $10$k updates but trails it on LIBERO-Plus. Extending training to $20$k keeps LIBERO success at $97.45\%$ and raises LIBERO-Plus success from $48.47\%$ to $55.58\%$.

\begin{table}[ht]
\vspace{-2mm}
\caption{Success rates in percent for FastWAM and \modelname{} with Wan at two training budgets. Each run has its own full learning-rate schedule, so the $10$k rows are independently trained endpoints.}
\label{tab:budget}
\begin{center}
\small\setlength{\tabcolsep}{7pt}
\begin{tabular}{lccc}
\toprule
Policy & Updates & LIBERO & LIBERO-Plus \\
\midrule
\multirow{2}{*}{FastWAM} & $10$k & 95.90 & 55.25 \\
 & $20$k & 97.00 & 55.31 \\
\midrule
\multirow{2}{*}{\modelname{} (Wan)} & $10$k & \textbf{97.45} & 48.47 \\
 & $20$k & \textbf{97.45} & \textbf{55.58} \\
\bottomrule
\end{tabular}
\end{center}
\vspace{-2mm}
\end{table}

\FloatBarrier

\section{Additional evaluations and analyses}
\label{app:inprogress}

\subsection{Backbone and target initialization}
\label{app:initfull}

Table~\ref{tab:factorialfull} expands Table~\ref{tab:factorial} using the Wan settings of Appendix~\ref{app:config}. The action expert, policy VAE, and text conditioner remain pretrained in every configuration.

\begin{table}[htbp]
\vspace{-2mm}
\caption{LIBERO success rates in percent for pretrained and random initialization of the Wan backbone and future anchor.}
\vspace{2mm}
\label{tab:factorialfull}
\centering
\small\setlength{\tabcolsep}{5pt}
\begin{tabular}{@{}llccccc@{}}
\toprule
Backbone & Anchor & Spatial & Object & Goal & Long & AVG \\
\midrule
Pretrained & Pretrained & \textbf{98.2} & \textbf{100.0} & \textbf{97.6} & 94.0 & 97.45 \\
Pretrained & Random & 98.0 & 99.8 & 97.0 & \textbf{95.8} & \textbf{97.65} \\
Random & Random & 85.6 & 98.0 & 88.2 & 65.8 & 84.40 \\
Random & Pretrained & 76.0 & 95.0 & 70.4 & 43.2 & 71.15 \\
\bottomrule
\end{tabular}
\vspace{-2mm}
\end{table}

\subsection{LIBERO and LIBERO-Plus results}
\label{app:liberoplus}

Table~\ref{tab:libero} reports success rates on LIBERO's Spatial, Object, Goal, and Long suites, with AVG computed as their mean. LIBERO-Plus perturbs camera pose, robot state, language, lighting, background, sensor noise, and layout \citep{liberoplus2025}. Using ImageWAM's evaluator, we sample $15\%$ of each suite \citep{imagewam2026}: 361 Spatial, 378 Object, 389 Goal, and 378 Long tasks. All local policies share these $1{,}506$ tasks with one rollout per task. Overall success is micro-averaged across tasks. Published scores in Table~\ref{tab:mainplus} are from \citet{liberoplus2025}.

\begin{figure}[!htbp]
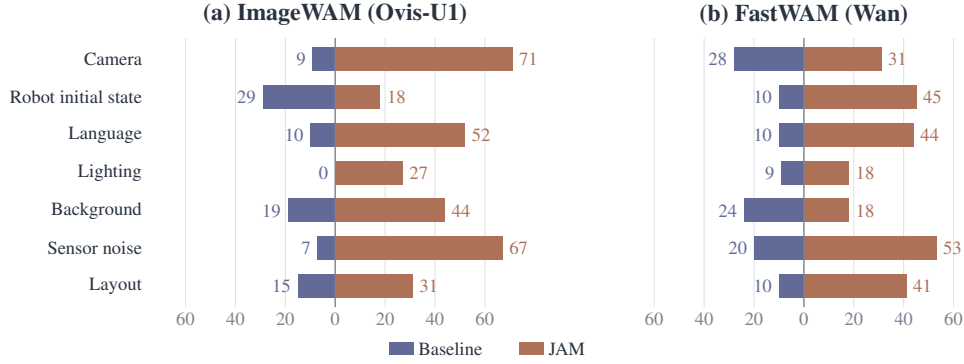

\centering
\input{Figures/data/liberoplus_pairs.tex}
\input{Figures/data/liberoplus_pairs_fastwam.tex}
\definecolor{pairbase}{HTML}{626B98}
\definecolor{pairjam}{HTML}{AD7158}
\definecolor{pairink}{HTML}{293340}
\definecolor{pairmuted}{HTML}{7A8590}
\definecolor{pairgrid}{HTML}{DCE1E6}
\begin{tikzpicture}[x=1cm,y=1cm,font=\footnotesize]
  \tikzset{
    paneltitle/.style={font=\small\bfseries,anchor=south,text=pairink,inner sep=0pt},
    ticklabel/.style={font=\scriptsize,text=pairmuted,anchor=north,inner sep=0pt},
    catlabel/.style={font=\scriptsize,anchor=east,text=pairink,inner sep=0pt},
    value/.style={font=\scriptsize,inner sep=1pt},
  }
  \def\rowA{3.6}\def\rowB{3.1}\def\rowC{2.6}\def\rowD{2.1}\def\rowE{1.6}\def\rowF{1.1}\def\rowG{0.6}
  \def\ylo{0.32}\def\yhi{3.88}

  \foreach \cx in {4.95,11.15}{
    \foreach \dx/\tick in {0.0/0,0.66/20,1.32/40,1.98/60}{
      \draw[pairgrid,line width=0.35pt] ({\cx-\dx},\ylo)--({\cx-\dx},\yhi);
      \draw[pairgrid,line width=0.35pt] ({\cx+\dx},\ylo)--({\cx+\dx},\yhi);
      \node[ticklabel] at ({\cx-\dx},{\ylo-0.06}) {\tick};
      \ifdim\dx pt>0pt
        \node[ticklabel] at ({\cx+\dx},{\ylo-0.06}) {\tick};
      \fi
    }
  }

  \newcommand{\pairbar}[4]{%
    \pgfmathsetmacro{\leftbar}{0.033*#3}%
    \pgfmathsetmacro{\rightbar}{0.033*#4}%
    \fill[pairbase] ({#2-\leftbar},{#1-0.155}) rectangle (#2,{#1+0.155});
    \fill[pairjam] (#2,{#1-0.155}) rectangle ({#2+\rightbar},{#1+0.155});
    \node[value,anchor=east,text=pairbase] at ({#2-\leftbar-0.05},#1) {#3};
    \node[value,anchor=west,text=pairjam] at ({#2+\rightbar+0.05},#1) {#4};
  }

  \foreach \y/\lab in {\rowA/Camera,\rowB/{Robot initial state},\rowC/Language,\rowD/Lighting,\rowE/Background,\rowF/{Sensor noise},\rowG/Layout}{
    \node[catlabel] at (2.4,\y) {\lab};
  }

  \pairbar{\rowA}{4.95}{\PairCameraBaselineOnly}{\PairCameraCandidateOnly}
  \pairbar{\rowB}{4.95}{\PairRobotBaselineOnly}{\PairRobotCandidateOnly}
  \pairbar{\rowC}{4.95}{\PairLanguageBaselineOnly}{\PairLanguageCandidateOnly}
  \pairbar{\rowD}{4.95}{\PairLightBaselineOnly}{\PairLightCandidateOnly}
  \pairbar{\rowE}{4.95}{\PairBackgroundBaselineOnly}{\PairBackgroundCandidateOnly}
  \pairbar{\rowF}{4.95}{\PairNoiseBaselineOnly}{\PairNoiseCandidateOnly}
  \pairbar{\rowG}{4.95}{\PairLayoutBaselineOnly}{\PairLayoutCandidateOnly}

  \pairbar{\rowA}{11.15}{\FastPairCameraBaselineOnly}{\FastPairCameraCandidateOnly}
  \pairbar{\rowB}{11.15}{\FastPairRobotBaselineOnly}{\FastPairRobotCandidateOnly}
  \pairbar{\rowC}{11.15}{\FastPairLanguageBaselineOnly}{\FastPairLanguageCandidateOnly}
  \pairbar{\rowD}{11.15}{\FastPairLightBaselineOnly}{\FastPairLightCandidateOnly}
  \pairbar{\rowE}{11.15}{\FastPairBackgroundBaselineOnly}{\FastPairBackgroundCandidateOnly}
  \pairbar{\rowF}{11.15}{\FastPairNoiseBaselineOnly}{\FastPairNoiseCandidateOnly}
  \pairbar{\rowG}{11.15}{\FastPairLayoutBaselineOnly}{\FastPairLayoutCandidateOnly}

  \foreach \cx in {4.95,11.15}{
    \draw[pairmuted,line width=0.5pt] (\cx,\ylo)--(\cx,\yhi);
  }

  \node[paneltitle] at (4.95,4.05) {(a) ImageWAM (Ovis-U1)};
  \node[paneltitle] at (11.15,4.05) {(b) FastWAM (Wan)};

  \node[font=\scriptsize,text=pairink,anchor=north,inner sep=0pt] at ([yshift=-0.22cm]current bounding box.south)
    {\raisebox{0.3pt}{\textcolor{pairbase}{\rule{0.32cm}{0.16cm}}}\;Baseline\qquad\raisebox{0.3pt}{\textcolor{pairjam}{\rule{0.32cm}{0.16cm}}}\;\modelname{}};
\end{tikzpicture}%
\caption{Paired LIBERO-Plus outcomes on \PairTaskCount{} shared tasks. Left bars count tasks solved only by the baseline; right bars count tasks solved only by \modelname{}. Panel a uses Ovis-U1, and panel b uses Wan with VAE-direct $[1,2]$ targets.}
\label{fig:robustnesspairs}
\end{figure}

Figure~\ref{fig:robustnesspairs} compares paired outcomes for Ovis-U1 in the left panel and Wan in the right. Ovis-U1 gains most under camera and sensor-noise perturbations, with 71 \modelname{}-only versus 9 baseline-only camera successes. Wan gains most under robot-state perturbations, with 45 \modelname{}-only versus 10 baseline-only successes; its camera outcomes are nearly balanced at 31 versus 28. Both improve in six of seven categories. The exceptions are robot-state perturbations for Ovis-U1 and background perturbations for Wan.

\begin{table}[htbp]
\vspace{-2mm}
\caption{LIBERO success rates in percent by suite. AVG is the four-suite mean. Published rows follow FastWAM~\citep{fastwam2026}, except GR00T N1, reported by LingBot-VA~\citep{lingbotva2026}; they retain their source protocols.}
\vspace{2mm}
\label{tab:libero}
\centering
\small
\setlength{\tabcolsep}{2.7pt}
\begin{tabular}{@{}l c ccccc@{}}
\toprule
Method & P.T. & Spatial & Object & Goal & Long & AVG \\
\midrule
\multicolumn{7}{@{}l}{\emph{VLM-based policies}} \\
OpenVLA~\citep{openvla2024} & \checkmark & 84.7 & 88.4 & 79.2 & 53.7 & 76.5 \\
GR00T N1~\citep{gr00tn1_2025} & \checkmark & 94.4 & 97.6 & 93.0 & 90.6 & 93.9 \\
$\pi_0$~\citep{pi0_2024} & \checkmark & 96.8 & 98.8 & 95.8 & 85.2 & 94.1 \\
$\pi_{0.5}$~\citep{pi05_2025} & \checkmark & \textbf{98.8} & 98.2 & \textbf{98.0} & \textbf{92.4} & \textbf{96.9} \\
\modelname{} (Qwen3.5-2B, Qwen3.5 targets)$^\dagger$ & $\times$ & 97.8 & \textbf{99.6} & 97.2 & 91.6 & 96.55 \\
\midrule
\multicolumn{7}{@{}l}{\emph{Video generator-based policies}} \\
Motus~\citep{motus2026} & \checkmark & 96.8 & 99.8 & 96.6 & 97.6 & 97.7 \\
LingBot-VA~\citep{lingbotva2026} & \checkmark & 98.5 & 99.6 & 97.2 & \textbf{98.5} & \textbf{98.5} \\
FastWAM$^\dagger$~\citep{fastwam2026} & $\times$ & 97.4 & 99.6 & 96.6 & 94.4 & 97.00 \\
\modelname{} (Wan, VAE targets)$^\dagger$ & $\times$ & \textbf{98.8} & \textbf{100.0} & \textbf{98.0} & 95.8 & 98.15 \\
\midrule
\multicolumn{7}{@{}l}{\emph{Image generator-based policies}} \\
ImageWAM (Ovis-U1)$^\dagger$~\citep{imagewam2026} & $\times$ & 97.8 & \textbf{100.0} & 93.6 & 80.2 & 92.90 \\
\modelname{} (Ovis-U1, Yak targets)$^\dagger$ & $\times$ & \textbf{98.6} & 99.6 & \textbf{98.4} & \textbf{90.4} & \textbf{96.75} \\
\bottomrule
\end{tabular}
\vspace{-2mm}
\end{table}

\FloatBarrier

\FloatBarrier

\subsection{RoboTwin 2.0 evaluation}
\label{app:robotwin}

For the local RoboTwin 2.0 results in Table~\ref{tab:robotwinmain}, we follow FastWAM and ImageWAM \citep{fastwam2026,imagewam2026}. We evaluate 50 tasks with 100 trials per task and setting, using unseen instructions and SAPIEN 3.0.0b1. Clean uses the default scene; Randomized varies tabletop clutter, background texture, lighting, table height, and instruction wording.
\begin{table}[htbp]
\vspace{-2mm}
\caption{RoboTwin 2.0 success rates in percent. AVG is the mean of the Clean and Randomized settings. Published Clean and Randomized values are from FastWAM~\citep{fastwam2026}, Table~1; rows marked $\dagger$ are evaluated by us.}
\vspace{1mm}
\label{tab:robotwinmain}
\centering
\small
\setlength{\tabcolsep}{5.0pt}
\begin{tabular}{@{}l c ccc@{}}
\toprule
Method & P.T. & Clean & Rand. & AVG \\
\midrule
\multicolumn{5}{@{}l}{\emph{VLM-based policies}} \\
$\pi_0$~\citep{pi0_2024} & \checkmark & 65.92 & 58.40 & 62.16 \\
$\pi_{0.5}$~\citep{pi05_2025} & \checkmark & 82.74 & 76.76 & 79.75 \\
\modelname{} (Qwen3.5-2B)$^\dagger$ & $\times$ & 89.70 & 89.86 & \textbf{89.78} \\
\midrule
\multicolumn{5}{@{}l}{\emph{Video generator-based policies}} \\
Motus~\citep{motus2026} & \checkmark & 88.66 & 87.02 & 87.84 \\
LingBot-VA~\citep{lingbotva2026} & \checkmark & 92.90 & 91.50 & \textbf{92.20} \\
FastWAM (released weights)$^\dagger$ & $\times$ & 91.22 & 91.20 & 91.21 \\
\modelname{} (Wan)$^\dagger$ & $\times$ & 91.70 & 91.18 & 91.44 \\
\midrule
\multicolumn{5}{@{}l}{\emph{Image generator-based policies}} \\
ImageWAM (Ovis-U1, local training)$^\dagger$ & $\times$ & 89.14 & 88.78 & 88.96 \\
ImageWAM (FLUX.2, released weights)$^\dagger$ & $\times$ & 92.84 & 93.04 & 92.94 \\
\modelname{} (Ovis-U1)$^\dagger$ & $\times$ & 90.34 & 89.46 & 89.90 \\
\modelname{} (FLUX.2)$^\dagger$ & $\times$ & 92.88 & 93.28 & \textbf{93.08} \\
\bottomrule
\end{tabular}
\vspace{-2mm}
\end{table}

\FloatBarrier

\subsection{Additional matched rollouts}
\label{app:morerollouts}

Panels a and b of Figure~\ref{fig:morerollouts} show LIBERO-Plus camera and layout perturbations; panels c and d show RoboTwin Clean and Randomized settings. Each LIBERO-Plus pair shares the initial state, and each RoboTwin pair shares the scene seed. Frame times are chosen separately for each policy to align manipulation stages.

\clearpage
\begingroup
\edef\jamRolloutsParskip{\the\parskip}
\noindent\begin{minipage}{\linewidth}
\setlength{\parskip}{\jamRolloutsParskip}
\setlength{\abovecaptionskip}{6pt}
\begingroup
\makeatletter
\def\@captype{figure}
\makeatother
\setlength{\parskip}{0pt}
\centering
\definecolor{wink}{HTML}{293340}\definecolor{wmuted}{HTML}{7A8590}\definecolor{wpanel}{HTML}{C8D4DE}
\resizebox{0.80\linewidth}{!}{%
\begin{tikzpicture}[x=1cm,y=1cm]
  \tikzset{
    tasktitle/.style={font=\scriptsize\bfseries,text=wink,anchor=south,inner sep=0pt},
    stage/.style={font=\scriptsize,text=wmuted,anchor=base,inner sep=0pt},
    method/.style={font=\scriptsize\bfseries,text=wink,rotate=90,anchor=center,inner sep=0pt},
  }
  \newcommand{\rollframe}[3]{\node[anchor=south west,inner sep=0pt] at (#1,#2) {\includegraphics[width=1.6cm]{#3}};}

  \node[method] at (0.20,6.765) {ImageWAM};
  \node[method] at (0.20,5.225) {\modelname{}};
  \node[tasktitle] at (3.755,7.86) {(a) Turn on the stove and put the moka pot on it};
  \foreach \k/\name in {0/Initial,1/Stove,2/Grasp,3/Terminal}{
    \node[stage] at ({0.42+\k*1.69+0.8},7.60) {\name};
    \rollframe{0.42+\k*1.69}{6.04}{Figures/qualitative/stove_native_\k.png}
    \rollframe{0.42+\k*1.69}{4.50}{Figures/qualitative/stove_mirai_\k.png}
  }
  \draw[wpanel,line width=0.6pt] (7.27,4.50)--(7.27,8.22);
  \node[tasktitle] at (10.785,7.86) {(b) Open the middle drawer of the cabinet};
  \foreach \k/\name in {0/Initial,1/Approach,2/Pull,3/Terminal}{
    \node[stage] at ({7.45+\k*1.69+0.8},7.60) {\name};
    \rollframe{7.45+\k*1.69}{6.04}{Figures/qualitative/cabinet_native_\k.png}
    \rollframe{7.45+\k*1.69}{4.50}{Figures/qualitative/cabinet_mirai_\k.png}
  }

  \node[method] at (0.20,2.365) {FastWAM};
  \node[method] at (0.20,0.825) {\modelname{}};
  \node[tasktitle] at (3.755,3.46) {(c) Put the can in the basket};
  \foreach \k/\name in {0/Initial,1/Grasp,2/Place,3/Terminal}{
    \node[stage] at ({0.42+\k*1.69+0.8},3.20) {\name};
    \rollframe{0.42+\k*1.69}{1.64}{Figures/qualitative/can_fastwam_\k.png}
    \rollframe{0.42+\k*1.69}{0.10}{Figures/qualitative/can_jam_\k.png}
  }
  \draw[wpanel,line width=0.6pt] (7.27,0.10)--(7.27,3.82);
  \node[tasktitle] at (10.785,3.46) {(d) Stack the red, green, and blue blocks};
  \foreach \k/\name in {0/Initial,1/First block,2/Second block,3/Terminal}{
    \node[stage] at ({7.45+\k*1.69+0.8},3.20) {\name};
    \rollframe{7.45+\k*1.69}{1.64}{Figures/qualitative/blocks_fastwam_\k.png}
    \rollframe{7.45+\k*1.69}{0.10}{Figures/qualitative/blocks_jam_\k.png}
  }
\end{tikzpicture}%
}
\caption{Matched rollouts with Ovis-U1 on LIBERO-Plus in panels a and b, and with Wan on RoboTwin 2.0 in panels c and d. ImageWAM fails to lift the moka pot in a and pulls the top drawer instead of the middle drawer in b. FastWAM knocks over the can in c and topples the stack in d. \modelname{} completes these selected tasks.}
\label{fig:morerollouts}
\par
\endgroup

\subsection{Action-to-visual attention}
\label{app:attention}

Using the deployed masks, we average attention over heads, 16 action queries, ten denoising steps, and captured layers. For each policy and view, we normalize maps to unit mass, subtract the mean over 100 reference frames from 20 LIBERO tasks, and retain positive values. We divide by the 99th percentile, clip to $[0,1]$, and apply power-$0.6$ compression. Difference maps show \modelname{} minus ImageWAM after this transformation on a $[-1,1]$ scale; magnitudes below $0.25$ appear gray.

In the Ovis-U1 examples of Figure~\ref{fig:attentionfull}, ImageWAM emphasizes the gripper and held object, while \modelname{} emphasizes the instructed placement target: the rack, plate, second plate, and microwave, in row order.

\medskip
\begingroup
\makeatletter
\def\@captype{figure}
\makeatother
\setlength{\parskip}{0pt}
\centering
\definecolor{wink}{HTML}{293340}\definecolor{wmuted}{HTML}{7A8590}
\resizebox{0.80\linewidth}{!}{%
\begin{tikzpicture}[x=1cm,y=1cm]
  \tikzset{
    colhead/.style={font=\tiny\bfseries,text=wink,anchor=south,inner sep=0pt},
    rowlab/.style={font=\scriptsize,text=wink,rotate=90,anchor=center,inner sep=0pt,align=center},
  }
  \newcommand{\cellsq}[3]{\node[anchor=south west,inner sep=0pt] at (#1,#2) {\includegraphics[width=2.05cm]{Figures/attention/cells6/#3}};}
  \foreach \c/\name in {0/Image,1/{ImageWAM (Ovis-U1)},2/{\modelname{} (Ovis-U1)},3/Difference,4/{\modelname{} (DINOv2-L)},5/{\modelname{} (Qwen3.5-2B)}}{
    \node[colhead] at ({0.75+\c*2.15+1.025},8.63) {\name};
  }
  \foreach \r/\pre/\lab in {0/S1/{LIBERO-Goal},1/S2/{LIBERO-Long},2/S3/{LIBERO-Plus\\camera},3/S4/{LIBERO-Plus\\layout}}{
    \node[rowlab] at (0.40,{6.47-\r*2.15+1.025}) {\lab};
    \foreach \c/\fam in {0/image,1/native,2/yak,3/diff,4/dinol,5/qwen}{
      \cellsq{0.75+\c*2.15}{6.47-\r*2.15}{\pre_agent_\fam.png}
    }
  }
  \draw[wmuted,line width=0.3pt,dash pattern=on 1.4pt off 1.4pt] (9.30,0.02) -- (9.30,9.25);
\end{tikzpicture}%
}
\caption{Action-to-visual attention in the agent view on LIBERO and LIBERO-Plus. Maps show attention above each policy's mean using the normalization in Appendix~\ref{app:attention}. Difference compares display intensities: red indicates higher values for \modelname{}, blue for ImageWAM.}
\label{fig:attentionfull}
\par
\endgroup

\end{minipage}\par
\endgroup
\clearpage

\subsection{Future-frame readout}
\label{app:readoutdiff}

We fit two per-token, two-layer MLP probes on each frozen Ovis-U1 policy to reconstruct the current frame and the frame $16$ steps ahead. Training and evaluation episodes are disjoint. The signed luminance difference between reconstructions gives the change maps in Figure~\ref{fig:readoutdiff}. The probes are used only for analysis.
In Figure~\ref{fig:readoutpsnr}, \modelname{} $[1]$ has the highest mean future-frame PSNR, slightly above ImageWAM and \modelname{} $[0]$. The ``Current'' baseline uses the current frame as its prediction of the future frame. Bracketed numbers denote the target offsets used during policy training.

\medskip
\noindent\begin{minipage}{\linewidth}
\begingroup
\vspace{-2mm}
\makeatletter
\def\@captype{figure}
\makeatother
\centering
\def\readoutPsnrX{0.80cm}
\begingroup%
\def\psnrCopy{16.13}%
\def\psnrImageWAM{19.43}%
\def\psnrOffsetZero{19.29}%
\def\psnrOffsetOne{19.57}%
\providecommand{\readoutPsnrX}{1.05cm}%
\newdimen\readoutPsnrY%
\readoutPsnrY=0.145cm\relax
\def\drawReadoutPsnr{%
\begin{tikzpicture}[x=\readoutPsnrX,y=\readoutPsnrY,font=\small]
  \definecolor{psnrInk}{HTML}{303943}
  \definecolor{psnrGrid}{HTML}{E5E9ED}
  \definecolor{psnrCopyColor}{HTML}{BDC5CD}
  \definecolor{psnrBaselineColor}{HTML}{6898B8}
  \definecolor{psnrOffsetZeroColor}{HTML}{E8B56D}
  \definecolor{psnrOffsetOneColor}{HTML}{C87942}

  \foreach \tick in {16,17,18,19,20} {
    \draw[psnrGrid,line width=0.4pt] (0,{4*(\tick-15)}) -- (7,{4*(\tick-15)});
  }
  \foreach \tick in {15,16,17,18,19,20} {
    \draw[psnrInk!65,line width=0.45pt] (-0.045,{4*(\tick-15)}) -- (0,{4*(\tick-15)});
    \node[anchor=east,text=psnrInk,inner sep=0pt,xshift=-4pt] at (0,{4*(\tick-15)}) {\tick};
  }
  \node[rotate=90,text=psnrInk] at (-0.83cm,11) {Avg. PSNR (dB) $\uparrow$};

  \foreach \pos/\value/\barcolor in {
    0.80/\psnrCopy/psnrCopyColor,
    2.60/\psnrImageWAM/psnrBaselineColor,
    4.40/\psnrOffsetZero/psnrOffsetZeroColor,
    6.20/\psnrOffsetOne/psnrOffsetOneColor} {
    \fill[\barcolor] (\pos-0.43,0) rectangle (\pos+0.43,{4*(\value-15)});
    \node[anchor=south,text=psnrInk,inner sep=0pt,yshift=4pt] at (\pos,{4*(\value-15)}) {\value};
  }

  \draw[psnrInk!65,line width=0.55pt] (0,22) -- (0,0) -- (7,0);
  \draw[psnrInk!65,line width=0.55pt] (0,0) -- ([yshift=-2pt]0,0);
  \draw[psnrInk!65,line width=0.65pt]
    ([xshift=-2.2pt,yshift=-4.2pt]0,0) -- ([xshift=2.2pt,yshift=-2pt]0,0)
    ([xshift=-2.2pt,yshift=-7.2pt]0,0) -- ([xshift=2.2pt,yshift=-5pt]0,0);
  \node[anchor=base,font=\fontsize{8}{9.5}\selectfont,inner sep=0pt,yshift=-16pt] at (0,0) {\strut};
  \foreach \pos/\name in {0.80/Current,2.60/ImageWAM,4.40/{\modelname{} $[0]$},6.20/{\modelname{} $[1]$}} {
    \node[anchor=base,text=psnrInk,font=\fontsize{8}{9.5}\selectfont,inner sep=0pt,yshift=-8pt] at (\pos,0) {\strut\name};
  }
\end{tikzpicture}%
}%
\ifdefined\readoutPsnrHeight
  \newsavebox{\readoutPsnrMeasureBox}%
  \sbox{\readoutPsnrMeasureBox}{\drawReadoutPsnr}%
  \readoutPsnrY=\dimexpr(\readoutPsnrHeight
    -\ht\readoutPsnrMeasureBox-\dp\readoutPsnrMeasureBox
    +22\readoutPsnrY)/22\relax
\fi
\drawReadoutPsnr%
\endgroup%

\vspace{-5mm}
\caption{Mean future-frame PSNR.}
\vspace{-1mm}
\label{fig:readoutpsnr}
\endgroup
\end{minipage}\par

\subsection{Statistical significance}
\label{app:evaldetail}

At fixed checkpoints, we compute 95\% percentile confidence intervals from $100{,}000$ task-bootstrap resamples. LIBERO has 40 tasks with 50 matched episodes each; LIBERO-Plus has $1{,}506$ tasks with one rollout each. All four intervals in Table~\ref{tab:pairedstats} favor \modelname{} and exclude zero. Supplementary exact two-sided McNemar tests use matched episodes for LIBERO and matched tasks for LIBERO-Plus; all $p<0.05$.

\begin{table}[htbp]
\vspace{-4mm}
\caption{Fixed-checkpoint comparisons with generative baselines, using VAE-direct $[1,2]$ targets for Wan and pretrained Yak $[1]$ for Ovis-U1. $\Delta$ is the gain in percentage points; counts are successes unique to each policy.}
\vspace{2mm}
\label{tab:pairedstats}
\centering
\footnotesize
\setlength{\tabcolsep}{4pt}
\begin{tabular}{@{}llrccc@{}}
\toprule
Policy family & Benchmark & $\Delta$ & 95\% task-bootstrap CI & Baseline only / \modelname{} only & McNemar $p$ \\
\midrule
FastWAM & LIBERO & $+1.15$ & $[+0.20,+2.10]$ & $30/53$ & $0.015$ \\
ImageWAM & LIBERO & $+3.85$ & $[+1.00,+7.40]$ & $42/119$ & $1.01\!\times\!10^{-9}$ \\
FastWAM & LIBERO-Plus & $+9.23$ & $[+6.84,+11.69]$ & $111/250$ & $1.91\!\times\!10^{-13}$ \\
ImageWAM & LIBERO-Plus & $+14.67$ & $[+12.15,+17.13]$ & $89/310$ & $9.69\!\times\!10^{-30}$ \\
\bottomrule
\end{tabular}
\vspace{-2mm}
\end{table}

\subsection{Independent training runs}
\label{app:multiseed}

Table~\ref{tab:multiseed} reports three training runs for each selected configuration: Wan with VAE-direct $[1,2]$ targets and Ovis-U1 with pretrained Yak $[1]$ targets. Following Appendix~\ref{app:impl}, we use $20$k updates for Wan and $21.7$k for Ovis-U1, both at global batch 128. Evaluation tasks and seeds remain fixed.
LIBERO-Plus success is $64.61\pm0.37\%$ for Wan and $79.92\pm0.43\%$ for Ovis-U1, indicating low variability across these three runs per configuration.

\begin{table}[htbp]
\vspace{-4mm}
\caption{Success rates in percent across three independent training runs. Summaries give mean $\pm$ sample standard deviation, computed before rounding.}
\vspace{2mm}
\label{tab:multiseed}
\centering
\small
\setlength{\tabcolsep}{2pt}
\renewcommand{\arraystretch}{1.04}
\newcommand{\seedstat}[2]{\makebox[4.4em][c]{$#1\!\pm\!#2$}}
\begin{tabular*}{\textwidth}{@{\extracolsep{\fill}}llcccccc@{}}
\toprule
Policy & Run & Spatial & Object & Goal & Long & AVG & LIBERO-Plus \\
\midrule
\multirow{4}{*}{\shortstack[l]{\modelname{}\\(Wan)}}
 & 1 & 98.80 & 100.00 & 98.00 & 95.80 & 98.15 & 64.54 \\
 & 2 & 98.60 & 99.80 & 98.20 & 95.40 & 98.00 & 64.28 \\
 & 3 & 99.00 & 100.00 & 97.80 & 96.20 & 98.25 & 65.01 \\
 & Mean $\pm$ SD & \seedstat{98.80}{0.20} & \seedstat{99.93}{0.12} & \seedstat{98.00}{0.20} & \seedstat{95.80}{0.40} & \seedstat{98.13}{0.13} & \seedstat{64.61}{0.37} \\
\midrule
\multirow{4}{*}{\shortstack[l]{\modelname{}\\(Ovis-U1)}}
 & 1 & 98.60 & 99.60 & 98.40 & 90.40 & 96.75 & 79.95 \\
 & 2 & 98.80 & 99.60 & 98.40 & 90.80 & 96.90 & 80.35 \\
 & 3 & 98.40 & 99.80 & 98.20 & 90.00 & 96.60 & 79.48 \\
 & Mean $\pm$ SD & \seedstat{98.60}{0.20} & \seedstat{99.67}{0.12} & \seedstat{98.33}{0.12} & \seedstat{90.40}{0.40} & \seedstat{96.75}{0.15} & \seedstat{79.92}{0.43} \\
\bottomrule
\end{tabular*}
\vspace{-2mm}
\end{table}

\FloatBarrier
\section{Extended related work}

\label{app:extendedrw}

Video-based policies and joint visual--action models combine action learning with visual prediction \citep{gr1_2024,gr2_2024,pad2024,uva2025}. VPP and Seer connect predictive visual features to control \citep{vpp2024,seer2024}. Other policies use generated observations or predictive latent context during action selection \citep{foreact2026,efficientwam2026,whentotrust2026,forewam2026}.

\paragraph{Action-conditioned latent prediction.}
V-JEPA 2-AC trains an action-conditioned predictor on robot trajectories using frozen video features, then uses latent rollouts for model-predictive control toward an image goal \citep{vjepa2_2025}. \modelname{} uses future targets to learn policy representations for imitation learning. It discards the prediction head after training and samples actions without latent rollouts or planning.

\paragraph{Predictive representations in action policies.}
FLARE mixes future, state, and action tokens in the action DiT~\citep{flare2025}. FRAPPE retains multiple prefix-and-LoRA streams at inference~\citep{frappe2026}. VLAFlow's MindWPI adds current-observation V-JEPA context accessible to action tokens~\citep{vlaflow2026}. \modelname{} supervises existing visual-backbone tokens through a separate training-only head, preserving the matched baseline's action computation.

\paragraph{Representations and visual generation.}
REPA aligns diffusion features with visual representations, AGRA adds semantic alignment to WAM co-training, and Video-Mirai adds foresight to autoregressive video diffusion; all retain visual generation \citep{repa2024,agra2026,videomirai2026}. WAM4D adds structured visual supervision to world-action learning \citep{wam4d2026}. \modelname{} removes visual generation during policy fine-tuning and uses no predicted future features as action inputs.

\paragraph{Frozen and adaptive targets.}
Frozen random encoders map input images to targets \citep{rnd2019,randomteachers2023}. Noise As Targets learns assignments to fixed random vectors \citep{noiseastargets2017}. EMA encoders update their weights during training \citep{byol2020}. Table~\ref{tab:teacherfull} compares frozen pretrained and EMA future anchors. Table~\ref{tab:factorial} and Appendix~\ref{app:initfull} vary the initialization of the policy and frozen future anchor.

\end{document}